\documentclass{article} 
\usepackage{iclr2027_conference,times}

\usepackage{amsmath,amsfonts,bm}

\def\eqref#1{equation~\ref{#1}}

\def\1{\bm{1}}

\DeclareMathAlphabet{\mathsfit}{\encodingdefault}{\sfdefault}{m}{sl}
\SetMathAlphabet{\mathsfit}{bold}{\encodingdefault}{\sfdefault}{bx}{n}

\usepackage{hyperref}
\usepackage{url}

\usepackage{multirow}
\usepackage{booktabs}
\usepackage{algorithm}
\usepackage{algpseudocode}
\algrenewcommand\algorithmicrequire{\textbf{Input:}}
\algrenewcommand\algorithmicensure{\textbf{Output:}}
\usepackage{graphicx}
\usepackage{subcaption}

\title{Graph-dependent pooling: adaptive granularity in hierarchical graph neural networks}

\author{Bohan Yoon \\
Institute Multiscale Matter and Systems \\
Ewha Womans University \\
Seoul, Republic of Korea \\
\texttt{i92run@ewha.ac.kr}
\And
Hyejin Jang \\
Division of AI Data Convergence \\
Hankuk University of Foreign Studies \\
Yongin-si, Gyeonggi-do, Republic of Korea \\
\texttt{hj@hufs.ac.kr}
}

\iclrfinalcopy 
\begin{document}

\maketitle

\begin{abstract}
Graph Neural Networks are widely used to learn structural representations of relational data. Hierarchical graph pooling learns multi-level representations by progressively coarsening graphs; however, prior methods rely on predefined cluster numbers or fixed pooling ratios, limiting adaptability to individual graphs. We propose Graph-Dependent Pooling (GDPool), which dynamically determines pooling granularity for each graph. We analyze optimization conflicts among layer-wise objectives through gradient alignment, showing that layer-wise objectives at different pooling levels can send conflicting gradients to shared parameters. GDPool uses self-attention to select graph-specific anchors representing structurally important nodes and constructs coarse graphs through soft node assignment. Experiments on metal-organic framework and molecular graph benchmarks show that GDPool consistently achieves the best predictive performance across all benchmarks. Evaluations using modularity and the effective number of clusters show that GDPool achieves consistently higher modularity than existing methods while the number of active clusters decreases monotonically with depth. Sensitivity analyses confirm stable behavior across hyperparameter and architectural choices. 
\end{abstract}

\section{Introduction}
\label{sec:Introduction}

Graph Neural Networks (GNNs) have become a widely used approach for learning from relational data, including molecules, crystal structures, and social networks \citep{gilmer2017mpnn,xu2019gin}. In molecular and materials property prediction, graph-based models represent atoms and bonds as graphs and directly predict graph-level properties \citep{xie2018cgcnn,chen2019megnet}. Effective graph-level prediction requires aggregating node-level information into a representation of the entire graph. While standard sum or mean readout aggregates all node representations in a single step, graph pooling progressively coarsens the graph to learn structural representations at multiple levels \citep{ying2018diffpool,grattarola2022understanding,liu2023graphpooling}. Hierarchical graph pooling has therefore emerged as a key approach for compressing complex graph structures while preserving salient information, with recent work also theoretically analyzing how pooling operations affect the expressive power of GNNs \citep{bianchi2023expressive}.

Hierarchical graph pooling involves two key design challenges. The first is determining how much to coarsen a graph. Clustering-based methods such as DiffPool, MinCutPool \citep{bianchi2020mincut}, and MemPool \citep{khasahmadi2020memory} typically require a predefined number of clusters, whereas selection-based methods such as Graph U-Nets \citep{gao2019graphunets}, SAGPool \citep{lee2019sagpool}, and ASAPool \citep{ranjan2020asap} rely on a predefined pooling ratio. To relax this fixed granularity, methods such as GMPool \citep{ko2023gmpool} and BN-Pool \citep{castellana2026bnpool} adapt the number of clusters to each input graph, but still require a hyperparameter specifying the maximum number of clusters. Moreover, learning graph-dependent coarsening consistently across multiple pooling levels remains an open challenge. The second challenge is how to learn these levels as a coherent hierarchy. Existing soft-assignment methods apply auxiliary objectives at individual levels \citep{ying2018diffpool,bianchi2020mincut}; however, because each higher-level graph is constructed from the output of the preceding pooling operation, its supervision target recursively depends on earlier assignments. Consequently, supervising each level only through its own local objective does not necessarily ensure structural consistency across the hierarchy.

To address these challenges, we propose Graph-Dependent Pooling (GDPool). GDPool identifies important nodes as graph-specific anchors and dynamically determines pooling granularity without a predefined number of clusters or pooling ratio. Self-attention estimates node importance, selects anchors, and softly assigns nodes to the selected anchors to construct a coarse graph. We further propose hierarchical supervision, which applies pooling objectives to the cumulative assignments so that every level is anchored to the original graph. GDPool thus learns graph-dependent hierarchical representations. We evaluate predictive performance and hierarchy quality using modularity \citep{newman2004finding} for structural cohesion, the effective number of clusters $K_{\mathrm{eff}}$ \citep{zheng2023online} for cluster utilization, and gradient alignment \citep{sener2018multi,yu2020gradient} among hierarchical objectives. The evaluation quantifies whether GDPool adapts to each graph while maintaining a consistent hierarchy. Our main contributions are as follows.
\begin{itemize}
    \item \textbf{Graph-dependent pooling.}
    We propose GDPool, which selects important nodes as anchors via self-attention and thereby determines pooling granularity without a predefined number of clusters or pooling ratio. Each cluster is anchored at an actual node, which makes the hierarchy directly interpretable.
    \item \textbf{Gradient conflict analysis and hierarchical supervision.}
    Measuring the gradient alignment between the auxiliary objectives of different pooling layers, we identify two sources of conflict: layer-wise supervision and a pooling budget that does not fit the graph. We mitigate the first by supervising the cumulative assignment of every level against the original graph, and the second by graph-dependent anchor selection.
    \item \textbf{Empirical validation across domains.}
    GDPool achieves the best MAE on all five targets of four MOF and molecular benchmarks against nine existing pooling methods, with a monotonically decreasing number of active clusters and consistently higher modularity. GDPool thus forms coherent hierarchies rather than merely reducing the number of nodes.
\end{itemize}

\section{Related work}
\label{sec:Related_Work}

\subsection{Hierarchical graph pooling}

Graph pooling can be grouped into three categories: clustering, node selection, and edge contraction \citep{grattarola2022understanding}. Clustering-based methods learn soft node-to-supernode assignments, including DiffPool, MinCutPool, and MemPool. TopKPool \citep{gao2019graphunets,cangea2018towards,NEURIPS2019_4c5bcfec}, SAGPool \citep{lee2019sagpool}, ASAPool, and MaxCutPool \citep{abate2025maxcutpool} select important nodes or local clusters, while EdgePool \citep{diehl2019edge} iteratively merges adjacent nodes. Clustering-based pooling retains explicit assignments of original nodes, enabling direct analysis of hierarchical structures and cluster-level properties.

In clustering-based pooling, selecting representative nodes and determining cluster numbers are key design choices. GAT \citep{velivckovic2017graph} and graph Transformers \citep{dwivedi2020generalization,ying2021transformers,kreuzer2021rethinking,rampavsek2022recipe} learn local and global node interactions through attention. While SAGPool and ASAPool rely on local scores and a fixed pooling ratio, GDPool scores nodes with graph-wide attention and lets the number of anchors follow each graph's attention distribution. Unlike ASAPool, which further restricts cluster membership to $h$-hop neighborhoods, GDPool assigns every node to all anchors without locality constraints or additional learnable parameters. More generally, prior pooling methods fix the number of clusters or the pooling ratio in advance, whereas clustering methods determine cluster numbers from data using eigengaps \citep{von2007tutorial}, nonparametric clustering \citep{kulis2011revisiting}, or modularity maximization \citep{newman2006modularity,blondel2008fast,traag2019louvain}. In graph pooling, HGP-SL \citep{zhang2019hierarchical} and SEP \citep{wu2022structural} consider adaptive structures, and GMPool \citep{ko2023gmpool} and BN-Pool also adapt cluster numbers to each graph. However, both must set a maximum number of clusters, and GMPool additionally requires $O(N^3)$ eigendecomposition at each level. After pooling, the coarse graph should preserve the original structure while supporting stable learning across multiple levels. Graph coarsening studies have explored multiresolution construction \citep{defferrard2016convolutional}, spectral and cut preservation \citep{loukas2019graph,loukas2018spectrally}, and learning-based coarsening and scalability \citep{cai2021graph,huang2021scaling,bacciu2023generalizing}. These coarsening methods maintain the composed mapping from the original graph to each level, whereas learned pooling methods define their clustering objectives on each coarsened graph separately; to our knowledge, applying these objectives to the composed assignment at every level has not been studied. Hierarchical GNNs such as Graph U-Net alternate message passing and pooling, while GDPool uses a pre-norm architecture \citep{xiong2020layer} to stabilize these levels.

\subsection{Deep supervision for hierarchical learning}

Hierarchical models derive effectiveness from a sequence of intermediate representations, making the supervision of intermediate levels an important design consideration. Deep supervision was introduced to improve the optimization and representation learning of deep networks by imposing auxiliary objectives directly on intermediate representations. Deeply-Supervised Nets \citep{lee2015deeply} attach auxiliary classifiers to hidden layers, and GoogLeNet \citep{szegedy2015going}, HED \citep{xie2015holistically}, and UNet++ \citep{zhou2018unet++} extend deep supervision to deeper and multi-scale architectures. Hierarchical graph pooling similarly employs layer-wise auxiliary objectives: DiffPool applies link prediction and entropy regularization to the adjacency and assignment at each level, and MinCutPool uses min-cut and orthogonality objectives. In both methods, each objective is defined with respect to the graph at the corresponding level, so the auxiliary objectives are defined locally at each level while sharing parameters.

When multiple auxiliary objectives jointly optimize shared parameters, the relationship among the gradients of the objectives becomes critical. Multi-task learning addresses gradient interactions by formulating training as multi-objective optimization \citep{sener2018multi}, adjusting gradient magnitudes or loss weights with GradNorm \citep{chen2018gradnorm}, applying uncertainty weighting \citep{kendall2018multi}, or directly modifying conflicting gradients with PCGrad \citep{yu2020gradient} and CAGrad \citep{liu2021conflict}. In particular, \citet{du2018adapting} weight auxiliary losses by the cosine similarity between auxiliary and main-task gradients, showing that auxiliary supervision is not always aligned with the task objective. However, gradient interactions among layer-wise auxiliary objectives in hierarchical graph pooling have received limited attention.

\section{Gradient interactions in hierarchical learning}
\label{sec:Gradient_Interactions_of_Hierarchical_Learning}

Hierarchical graph learning sequentially constructs graph representations at different levels from a single input graph. Layer-wise supervision assigns a separate objective to the representation at each pooling level, providing a direct learning signal for capturing meaningful information at each level. However, when objectives at different levels share parameters, the gradients of the objectives may point in different directions, so that an update improving the objective at one level degrades the objective at another level. We define the interference between layer-wise objectives as gradient conflict, quantified by the directional alignment of the gradients that the objectives send to shared parameters. Measuring gradient conflict during training, we examine whether layer-wise objectives, each defined on its own coarsened graph, are compatible with learning a structurally coherent hierarchy.

\subsection{Gradient conflict in hierarchical learning}
\label{sec:Gradient_Conflict_in_Hierarchical_Learning}

\begin{figure}[ht]
    \centering
    \begin{subfigure}[t]{0.63\linewidth}
        \centering
        \includegraphics[width=\linewidth,height=3.5cm]{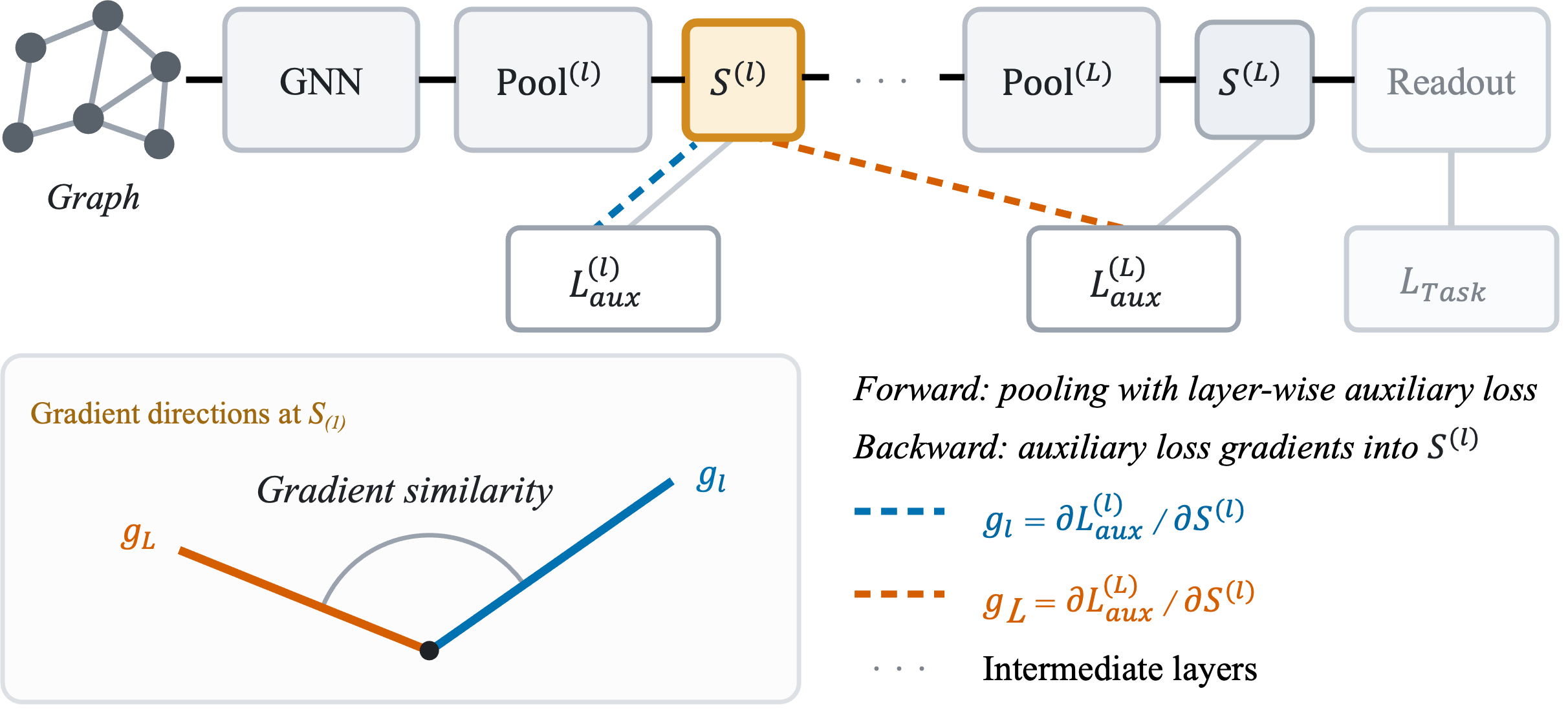}
        \caption{}
        \label{fig:gradient_similarity}
    \end{subfigure}
    \hfill
    \begin{subfigure}[t]{0.35\linewidth}
        \centering
        \includegraphics[width=\linewidth,height=3.5cm]{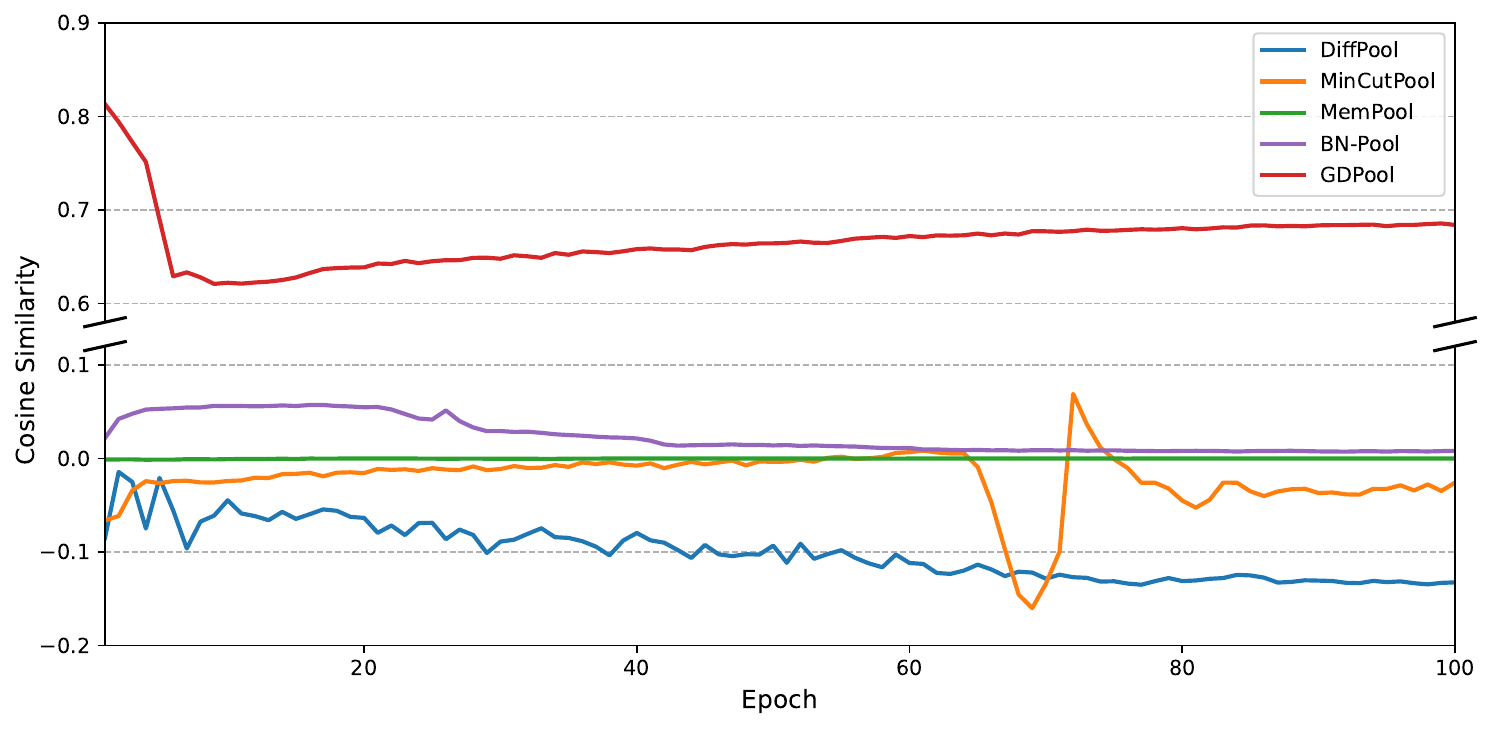}
        \caption{}
        \label{fig:gradient_similarity_base}
    \end{subfigure}
    \caption{(a) Cosine similarity used to quantify the alignment between gradients of layer-wise auxiliary objectives (b) Gradient alignment across hierarchy levels during training on the ARC-MOF dataset; the red line denotes GDPool.}
    \label{fig:gradient_conflict}
\end{figure}

Gradient conflict occurs when layer-wise objectives send gradients in opposing directions to shared parameters, so that improving one objective hinders another. To quantify gradient conflict, let $\mathcal{L}^{(l)}_{\mathrm{aux}}$ denote the auxiliary objective at the $l$-th pooling level. For two levels $l<L$, we take the gradients of both objectives with respect to the lower-level assignment $S^{(l)}$, $g_l=\partial\mathcal{L}^{(l)}_{\mathrm{aux}}/\partial S^{(l)}$ and $g_L=\partial\mathcal{L}^{(L)}_{\mathrm{aux}}/\partial S^{(l)}$ ($l=1$ in our experiments). Similar gradient directions at levels $i$ and $j$ indicate cooperative optimization, whereas opposing directions indicate that an update for one objective hinders the other. We quantify the directional relationship between two levels by gradient alignment. Positive alignment indicates that the two objectives favor similar update directions, whereas negative alignment signals likely gradient conflict; alignment near zero implies nearly orthogonal gradients with little mutual influence. Figure~\ref{fig:gradient_similarity} illustrates the measurement procedure, in which the gradients of the auxiliary objectives at each pooling level are computed and compared at $S^{(l)}$. Unlike a comparison of loss magnitudes, gradient alignment directly reveals the direction in which each layer-wise objective drives the shared parameters. Figure~\ref{fig:gradient_similarity_base} shows the gradient alignment between the auxiliary objectives of DiffPool, MinCutPool, MemPool, and BN-Pool, measured during training. The alignment of every baseline remains below 0.1 throughout training, either staying close to zero or falling below zero for extended periods. Layer-wise objectives therefore fail to provide cooperative learning signals, either conflicting with or barely influencing one another.

\subsection{Implications of gradient conflict}
\label{sec:Implications_of_gradient_conflict}

The observed gradient conflict suggests that applying a separate local objective at each level does not guarantee that the hierarchy is learned coherently. When layer-wise objectives send divergent gradients to shared parameters, learning at one level may remain incompatible with representation learning at another level, even if each objective is optimized individually. Satisfying the layer-wise objectives and preserving the structural relationships across the hierarchy can thus become distinct goals. Rather than assuming that layer-wise supervision alone yields a structurally coherent hierarchy, supervision should explicitly reflect the relationships across levels, since the representations at all levels are derived from a single original graph through successive coarsening. Building on the analysis, we propose hierarchical supervision, which applies existing pooling objectives to the cumulative assignments so that the supervision at each level reflects the structural relationships across levels. Section~\ref{sec:Gradient_Alignment_Analysis} analyzes the effect of hierarchical supervision on GDPool, further examining how gradient conflict across levels changes when a fixed pooling granularity is not enforced.

\section{Graph-dependent hierarchical pooling}
\label{sec:Graph-Dependent_Hierarchical_Pooling}

We propose GDPool, which adaptively determines pooling granularity according to the input graph. GDPool computes context-aware node importance by self-attention over the node representations and selects the most important nodes as graph-specific anchors. A coarse graph is then constructed through soft-assignment of each node to the selected anchors. GDPool thereby learns hierarchical representations whose granularity differs across graphs, without a predefined number of clusters or pooling ratio. 

\subsection{Context-aware anchor selection}
\label{sec:Context-Aware_Node_Attention}

GDPool computes context-aware node importance through self-attention, which accounts for interactions with the other nodes in the graph. Let $G=(X,A)$ denote an input graph, where $X\in\mathbb{R}^{N\times D}$ contains the $D$-dimensional representations of $N$ nodes and $A\in\mathbb{R}^{N\times N}$ is the adjacency matrix. GDPool uses the layer-normalized node representations as queries and keys, while retaining the original representations as values. GDPool computes the attention logits $R$ from the queries and keys, obtaining the attention map $H$ through row-wise softmax:
\begin{equation}
    H_{ij} = \frac{\exp(R_{ij})}{\sum_{k=1}^{N}\exp(R_{ik})}.
    \label{eq:attention_map}
\end{equation}
GDPool defines the context-aware importance of each node as the total attention the node receives from all query nodes. Important nodes are thus identified from contextual interactions across the entire graph rather than from independent node-wise scores. Nodes with importance above the mean $\bar{p}$ form the anchor set $\mathcal{A}$:
\begin{equation}
    p_j = \sum_i H_{ij}, \quad \mathcal{A} = \left\{ j \mid p_j > \bar{p} \right\}.
    \label{eq:anchor_selection}
\end{equation}
Since $H$ is row-stochastic, $\sum_j p_j = N$ and hence $\bar{p} = 1$; the anchor criterion thus selects nodes with $p_j > 1$ without a threshold hyperparameter. GDPool adds no learnable parameters beyond LayerNorm, since the queries, keys, and values are the node representations themselves. Because node importance depends on contextual interactions within each graph, the number of anchors varies across graphs, yielding graph-specific pooling granularity. Unlike prior adaptive pooling methods that require a predefined pooling ratio or an upper bound on the number of clusters, GDPool determines the anchor set directly from the graph-wide attention distribution. A node is selected exactly when it receives more than the uniform share of attention, so the granularity is induced by the importance distribution of each graph rather than fixed in advance or estimated as a separate budget. When no node exceeds the threshold, the node with the highest importance is included in $\mathcal{A}$ to guarantee at least one anchor.

\subsection{Graph-dependent assignment and coarsening}
\label{sec:Graph-Dependent_Assignment}

Based on the selected graph-specific anchors, GDPool constructs a hierarchical representation through soft-assignment of each node to the anchors. Specifically, the assignment score $R^{\mathcal{A}}_{ik}$ between the $i$-th node and the $k$-th anchor $a_k \in \mathcal{A}$ is given by the attention score $R_{a_k i}$. Normalizing the assignment scores over the $C$ selected anchors yields the soft-assignment matrix $S\in\mathbb{R}^{N\times C}$:
\begin{equation}
    S_{ik} = \frac{\exp(R^{\mathcal{A}}_{ik})}{\sum_{l=1}^{C}\exp(R^{\mathcal{A}}_{il})}.
    \label{eq:soft_assignment}
\end{equation}
Each node therefore holds continuous assignment weights over all selected anchors rather than belonging to a single anchor. The weight $S_{ik}$ indicates how strongly anchor $a_k$ represents node $i$. Using $S$, GDPool coarsens the input graph into a coarse graph $G'=(X',A')$ with one coarse node per anchor. The coarse node representations and adjacency matrix are given by $X' = S^\top X \in\mathbb{R}^{C\times D}$ and $A' = S^\top A S \in\mathbb{R}^{C\times C}$. Each coarse node aggregates the representations of the original nodes according to the assignment weights, so that, unlike hard clustering, an original node can contribute to multiple coarse nodes. The detailed procedure of GDPool is provided in Appendix~\ref{sec:appendix_gdpool_algorithm}.

\subsection{Hierarchical architecture}
\label{sec:Hierarchical_Architecture}

\begin{figure}[ht]
\centering
\includegraphics[width=0.7\linewidth]{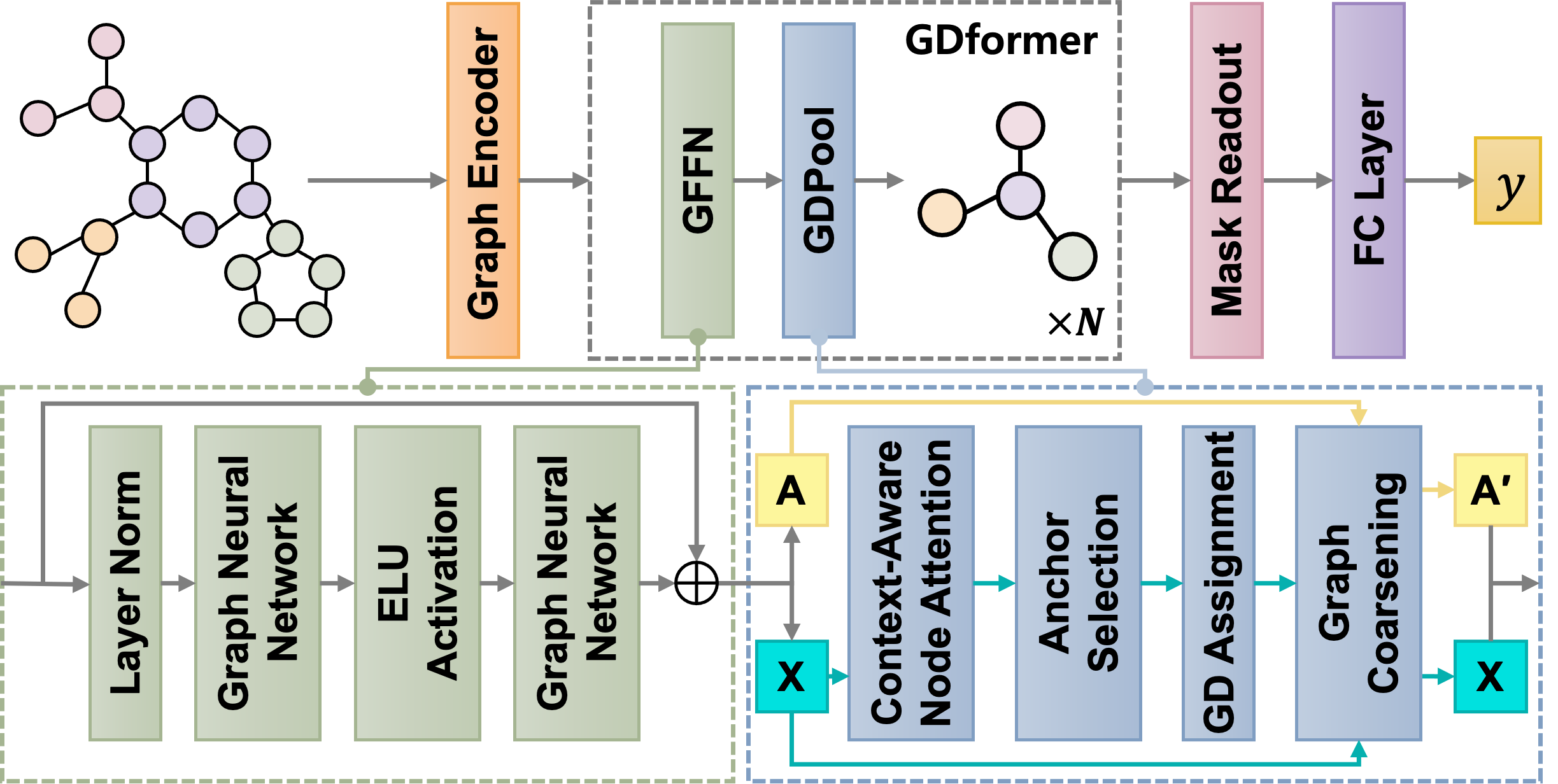}
\caption{Overview of the proposed hierarchical graph learning architecture, which progressively learns hierarchical graph representations using GDPool.}
\label{fig:framework}
\end{figure}

As shown in Figure~\ref{fig:framework}, GDPool is integrated into a hierarchical architecture that stacks GDformer blocks on a graph encoder, each block consisting of a Graph Feed-Forward Network (GFFN) followed by GDPool. Given an input graph $G^{(l)}$, the GFFN updates the node representations of $G^{(l)}$, after which GDPool constructs the coarse graph $G^{(l+1)}$ as the input to the next level. Repeated application of GDformer blocks yields hierarchical graph representations with graph-dependent pooling granularity. Because graphs within a mini-batch differ in size, the graphs at each level are padded to a common size, with a mask excluding padded nodes from attention, assignment, and readout. Masked readouts from all levels are aggregated for graph-level prediction. Appendix~\ref{sec:appendix_gdpool_layer} illustrates the GDPool layer.

To preserve the structural information of the original graph at each level, we adopt the auxiliary link prediction objective $\mathcal{L}_{\mathrm{Link}}$ and the entropy regularization $\mathcal{L}_{\mathrm{Ent}}$ of DiffPool. Under layer-wise supervision, existing pooling methods learn structural consistency between the input and pooled graphs using the assignment matrix of the current level. In hierarchical pooling, however, the original nodes are connected to the coarse nodes at higher levels through a chain of assignment matrices. Hierarchical supervision therefore defines the objective at each level on the assignment accumulated along the hierarchy rather than on the assignment of the corresponding level alone.

Based on the accumulated assignment, we apply link prediction and entropy regularization at each level so that the hierarchical representations maintain the structural relationships of the original graph. The overall hierarchical supervision is therefore
\begin{equation}
\mathcal{L}_{Hier}=\sum_{l=1}^{L}\left(\mathcal{L}_{\mathrm{Link}}(S_{\mathrm{Hier}}^{(l)})+\lambda\mathcal{L}_{\mathrm{Ent}}(S_{\mathrm{Hier}}^{(l)})\right).
    \label{eq:total_hierarchical_supervision}
\end{equation}
The accumulated assignment from the first to the $l$-th level is $S_{\mathrm{Hier}}^{(l)} = S^{(1)}S^{(2)}\cdots S^{(l)}$, with $\lambda$ weighting the entropy regularization. Each level is thus supervised against the original graph rather than only its own coarsened graph, which provides a common reference across levels in contrast to the layer-wise objectives of Section~\ref{sec:Gradient_Interactions_of_Hierarchical_Learning}. Since $S^{(l)}_{\mathrm{Hier}}$ is row-stochastic like $S^{(l)}$, the objectives are applied without modification, and the first-level term coincides with layer-wise supervision, so the two schemes differ only in the terms above the first level. The final training objective is the sum of the prediction loss and the hierarchical supervision. A detailed complexity comparison between GDPool and existing pooling methods is provided in Appendix~\ref{sec:appendix_complexity}.

\section{Experiments}
\label{sec:Experiments}

We conduct a comprehensive evaluation to assess the effectiveness of GDPool in graph property prediction and hierarchical graph representation learning. For a fair comparison, all pooling methods within each dataset share the same encoder, GFFN, and training settings, so that only the pooling operator differs. On the MOF benchmarks, graphs constructed from 3D crystal structures are encoded by MMAGNN~\citep{teng2025atomic}, with SAGEConv~\citep{hamilton2017inductive} used in each GFFN. On the molecular benchmarks, the graph encoder consists of four GINEConv~\citep{hu2020pretraining} layers to exploit the discrete bond-type edge features provided by the datasets, with GINEConv also used in each GFFN. We set the GFFN expansion ratio to 2 and the entropy weight $\lambda$ to 0.05, whose effects are further analyzed in Section~\ref{sec:Sensitivity_to_Model_Components}. The details of the architectures used in each experiment and the hyperparameter selection procedure are described in Appendix~\ref{sec:appendix_model_configurations}.

\paragraph{Datasets.}
We evaluate GDPool on four graph regression datasets spanning MOFs and molecular graphs: ARC-MOF~\citep{burner2023arc}, hMOF~\citep{wilmer2012large}, ZINC~\citep{dwivedi2023benchmarking}, and QM9~\citep{ramakrishnan2014quantum}. ARC-MOF and hMOF contain relatively large material graphs and are used for gas adsorption prediction, while ZINC and QM9 contain smaller molecular graphs and are used for solubility and dipole moment prediction, respectively. Dataset statistics are summarized in Table~\ref{tab:dataset_statistics}. All targets are used in their native units without normalization, and both molecular benchmarks are treated as 2D graphs without atomic coordinates; reported errors are therefore absolute and comparable only across pooling operators, not to geometry-based models (details in Appendix~\ref{sec:appendix_datasets_details}).

\paragraph{Baseline methods.}
\label{sec:baseline_methods}

We compare GDPool with hierarchical pooling methods from the three categories in Section~\ref{sec:Related_Work}: \textbf{clustering-based} (DiffPool, MinCutPool, MemPool, and BN-Pool), \textbf{node selection-based} (TopKPool, SAGPool, ASAPool, and MaxCutPool), and \textbf{edge contraction-based} (EdgePool) pooling. Except for BN-Pool, which infers the number of active clusters within a specified maximum, clustering-based methods require a predefined number of clusters. Node selection-based methods mainly coarsen graphs according to a predefined pooling ratio.

\subsection{Comparison with existing pooling methods}
\label{sec:Comparison_with_Existing_Pooling_Methods}

To validate the performance of GDPool, we use the four benchmark datasets. The targets are the CO$_2$ uptake at 0.15 bar on ARC-MOF, the CO$_2$ uptake at 0.1 bar and the CH$_4$ uptake at 0.9 bar on hMOF, the constrained solubility on ZINC, and the dipole moment on QM9. Predictive performance is evaluated with Mean Absolute Error (MAE), Root Mean Squared Error (RMSE), and $R^2$, where lower MAE and RMSE and higher $R^2$ indicate better prediction. As shown in Table~\ref{tab:performance_comparison}, GDPool achieves the best MAE on every prediction task across the datasets compared. GDPool therefore improves consistently on MOF and molecular graphs with different characteristics, rather than on a particular dataset or property. These results indicate that the anchor-based pooling of GDPool is effective for graph representation learning across graph types. The same trend holds for RMSE and $R^2$ (Tables~\ref{tab:performance_comparison_rmse} and~\ref{tab:performance_comparison_r2}). ASAPool and MaxCutPool did not train stably in our setting; we report their results as obtained and discuss this in Appendix~\ref{sec:appendix_additional_property_prediction}.
\begin{table}
\caption{MAE of graph pooling methods on the benchmark datasets (mean $\pm$ std over three seeds; $\pm0.000$ denotes std below 0.0005; units in Appendix~\ref{sec:appendix_datasets_details}). The best results are shown in bold.}
\label{tab:performance_comparison}
\begin{center}
\begin{tabular}{lccccc}
\toprule
\multirow{2}{*}{\textbf{Method}} & \multicolumn{1}{c}{\textbf{ARC-MOF}} & \multicolumn{2}{c}{\textbf{hMOF}} & \multicolumn{1}{c}{\textbf{ZINC}} & \multicolumn{1}{c}{\textbf{QM9}} \\
\cmidrule(lr){2-2}\cmidrule(lr){3-4}\cmidrule(lr){5-5}\cmidrule(lr){6-6}
& \multicolumn{1}{c}{CO$_2$@0.15} & \multicolumn{1}{c}{CO$_2$@0.1} & \multicolumn{1}{c}{CH$_4$@0.9} & \multicolumn{1}{c}{Solubility} & \multicolumn{1}{c}{Dipole Moment} \\
\midrule
DiffPool
& $0.184 {\scriptstyle \pm 0.014}$ & $0.159 {\scriptstyle \pm 0.002}$ 
& $0.137 {\scriptstyle \pm 0.010}$ & $0.323 {\scriptstyle \pm 0.007}$ 
& $0.483 {\scriptstyle \pm 0.012}$ \\
MinCutPool
& $0.158 {\scriptstyle \pm 0.001}$ & $0.129 {\scriptstyle \pm 0.001}$ 
& $0.110 {\scriptstyle \pm 0.002}$ & $0.319 {\scriptstyle \pm 0.002}$ 
& $0.452 {\scriptstyle \pm 0.003}$ \\
MemPool
& $0.157 {\scriptstyle \pm 0.000}$ & $0.129 {\scriptstyle \pm 0.000}$ 
& $0.110 {\scriptstyle \pm 0.000}$ & $0.316 {\scriptstyle \pm 0.002}$ 
& $0.448 {\scriptstyle \pm 0.002}$ \\
BN-Pool
& $0.160 {\scriptstyle \pm 0.004}$ & $0.130 {\scriptstyle \pm 0.001}$ 
& $0.113 {\scriptstyle \pm 0.001}$ & $0.322 {\scriptstyle \pm 0.003}$ 
& $0.460 {\scriptstyle \pm 0.000}$ \\
TopKPool
& $0.175 {\scriptstyle \pm 0.004}$ & $0.137 {\scriptstyle \pm 0.002}$ 
& $0.117 {\scriptstyle \pm 0.002}$ & $0.470 {\scriptstyle \pm 0.039}$ 
& $0.469 {\scriptstyle \pm 0.015}$ \\
SAGPool
& $0.167 {\scriptstyle \pm 0.002}$ & $0.137 {\scriptstyle \pm 0.001}$ 
& $0.121 {\scriptstyle \pm 0.002}$ & $0.352 {\scriptstyle \pm 0.084}$ 
& $0.487 {\scriptstyle \pm 0.013}$ \\
ASAPool
& $0.406 {\scriptstyle \pm 0.416}$ & $0.422 {\scriptstyle \pm 0.000}$ 
& $0.477 {\scriptstyle \pm 0.000}$ & $0.741 {\scriptstyle \pm 0.064}$ 
& $0.502 {\scriptstyle \pm 0.000}$ \\
MaxCutPool
& $0.382 {\scriptstyle \pm 0.177}$ & $0.347 {\scriptstyle \pm 0.016}$ 
& $0.272 {\scriptstyle \pm 0.121}$ & $0.585 {\scriptstyle \pm 0.113}$ 
& $0.668 {\scriptstyle \pm 0.090}$ \\
EdgePool
& $0.161 {\scriptstyle \pm 0.001}$ & $0.131 {\scriptstyle \pm 0.000}$ 
& $0.113 {\scriptstyle \pm 0.002}$ & $0.320 {\scriptstyle \pm 0.002}$ 
& $0.452 {\scriptstyle \pm 0.002}$ \\
\midrule
\textbf{GDPool}
& $\textbf{0.151}{\scriptstyle \pm 0.000}$ & $\textbf{0.125}{\scriptstyle \pm 0.002}$
& $\textbf{0.106}{\scriptstyle \pm 0.001}$ & $\textbf{0.240}{\scriptstyle \pm 0.015}$
& $\textbf{0.402}{\scriptstyle \pm 0.003}$ \\
\bottomrule
\end{tabular}
\end{center}
\end{table}

\subsection{Hierarchical clustering behavior}
\label{sec:Hierarchical_Clustering_Behavior}

To analyze hierarchical graph representations, we evaluate clustering behavior according to the pooling process. We compare GDPool with DiffPool, MinCutPool, MemPool, and BN-Pool, which directly use soft assignment. We measure the structural cohesion of the assignment produced at each pooling level with modularity. As shown in Table~\ref{tab:modularity_keff}, the modularity of the baselines collapses once pooling levels are stacked, whereas GDPool maintains high modularity at every level. A near-uniform soft assignment gives $Q \approx 0$ by construction, so these values indicate assignments that do not follow the graph structure; such degenerate assignments have been noted previously \citep{bianchi2020mincut}. GDPool therefore aggregates nodes according to the structural relationships of the input graph rather than merely reducing the number of nodes. We further compare the nominal number of clusters with $K_{\mathrm{eff}}$, the number of clusters actually used during pooling. Natural hierarchical coarsening activates fewer clusters at higher levels as pooling proceeds. The baselines show no clear reduction: $K_{\mathrm{eff}}$ at the first level is smaller than at later levels, or most clusters remain active at every level, leaving little difference from the nominal number. Stacking pooling levels therefore does not necessarily reduce the granularity of the representations. In contrast, $K_{\mathrm{eff}}$ of GDPool decreases with depth, using more clusters at early levels to represent fine-grained structure while activating fewer clusters at later levels to form coarse-grained representations. GDPool thus forms an appropriate pooling granularity by selecting anchors from the node representations of each graph instead of enforcing a fixed number of clusters. Both metrics are defined in Appendix~\ref{sec:appendix_metrics}.
\begin{table}[ht]
\caption{Layer-wise Modularity and $K_{\mathrm{eff}}$ for CH$_4$@0.9 on the hMOF dataset.}
\label{tab:modularity_keff}
\begin{center}
\begin{tabular}{lcccccc}
\toprule
\multirow{2}{*}{\textbf{Method}} & \multicolumn{3}{c}{\textbf{Modularity}} & \multicolumn{3}{c}{$\boldsymbol{K}_{\mathrm{eff}}$} \\
\cmidrule(lr){2-4}\cmidrule(lr){5-7}
& Layer 1 & Layer 2 & Layer 3 & Layer 1 & Layer 2 & Layer 3 \\
\midrule
DiffPool
& $0.00 {\scriptstyle \pm 0.01}$ & $0.00 {\scriptstyle \pm 0.01}$ 
& $0.00 {\scriptstyle \pm 0.02}$ & $6.16 {\scriptstyle \pm 1.07}$  
& $2.21 {\scriptstyle \pm 0.45}$ & $2.76 {\scriptstyle \pm 0.53}$ \\
MinCutPool
& $0.10 {\scriptstyle \pm 0.08}$ & $0.02 {\scriptstyle \pm 0.03}$ 
& $0.01 {\scriptstyle \pm 0.01}$ & $3.67 {\scriptstyle \pm 2.42}$  
& $9.69 {\scriptstyle \pm 4.59}$ & $7.01 {\scriptstyle \pm 0.94}$ \\
MemPool
& $0.00 {\scriptstyle \pm 0.00}$ & $0.00 {\scriptstyle \pm 0.00}$ 
& $0.00 {\scriptstyle \pm 0.00}$ & $32.00 {\scriptstyle \pm 0.00}$  
& $16.00 {\scriptstyle \pm 0.00}$ & $8.00 {\scriptstyle \pm 0.00}$ \\
BN-Pool
& $0.00 {\scriptstyle \pm 0.00}$ & $0.00 {\scriptstyle \pm 0.00}$ 
& $0.00 {\scriptstyle \pm 0.00}$ & $31.88 {\scriptstyle \pm 0.01}$  
& $15.79 {\scriptstyle \pm 0.00}$ & $7.85 {\scriptstyle \pm 0.00}$ \\
\midrule
\textbf{GDPool}
& $0.39{\scriptstyle \pm 0.01}$ & $0.38{\scriptstyle \pm 0.00}$
& $0.21{\scriptstyle \pm 0.00}$ & $17.35{\scriptstyle \pm 1.03}$ 
& $9.53{\scriptstyle \pm 0.34}$ & $5.43{\scriptstyle \pm 0.09}$ \\
\bottomrule
\end{tabular}
\end{center}
\end{table}

\subsection{Graph-dependent anchor selection}
\label{sec:adaptive_anchor_selection}

\begin{figure}[ht]
    \centering
    \begin{subfigure}[t]{0.45\linewidth}
        \centering
        \includegraphics[width=\linewidth]{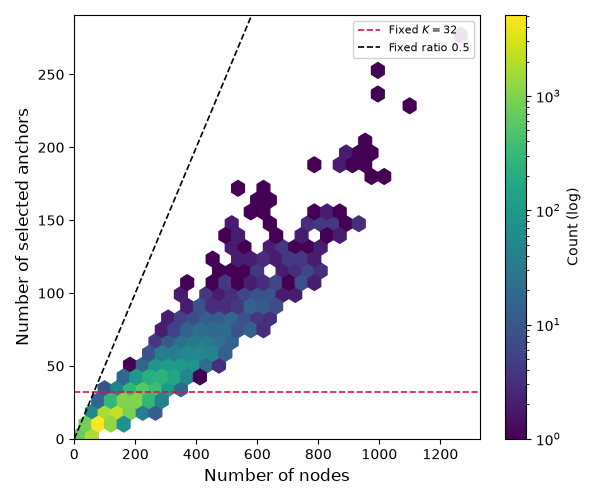}
        \caption{}
        \label{fig:node_anchor1}
    \end{subfigure}%
    \begin{subfigure}[t]{0.45\linewidth}
        \centering
        \includegraphics[width=\linewidth]{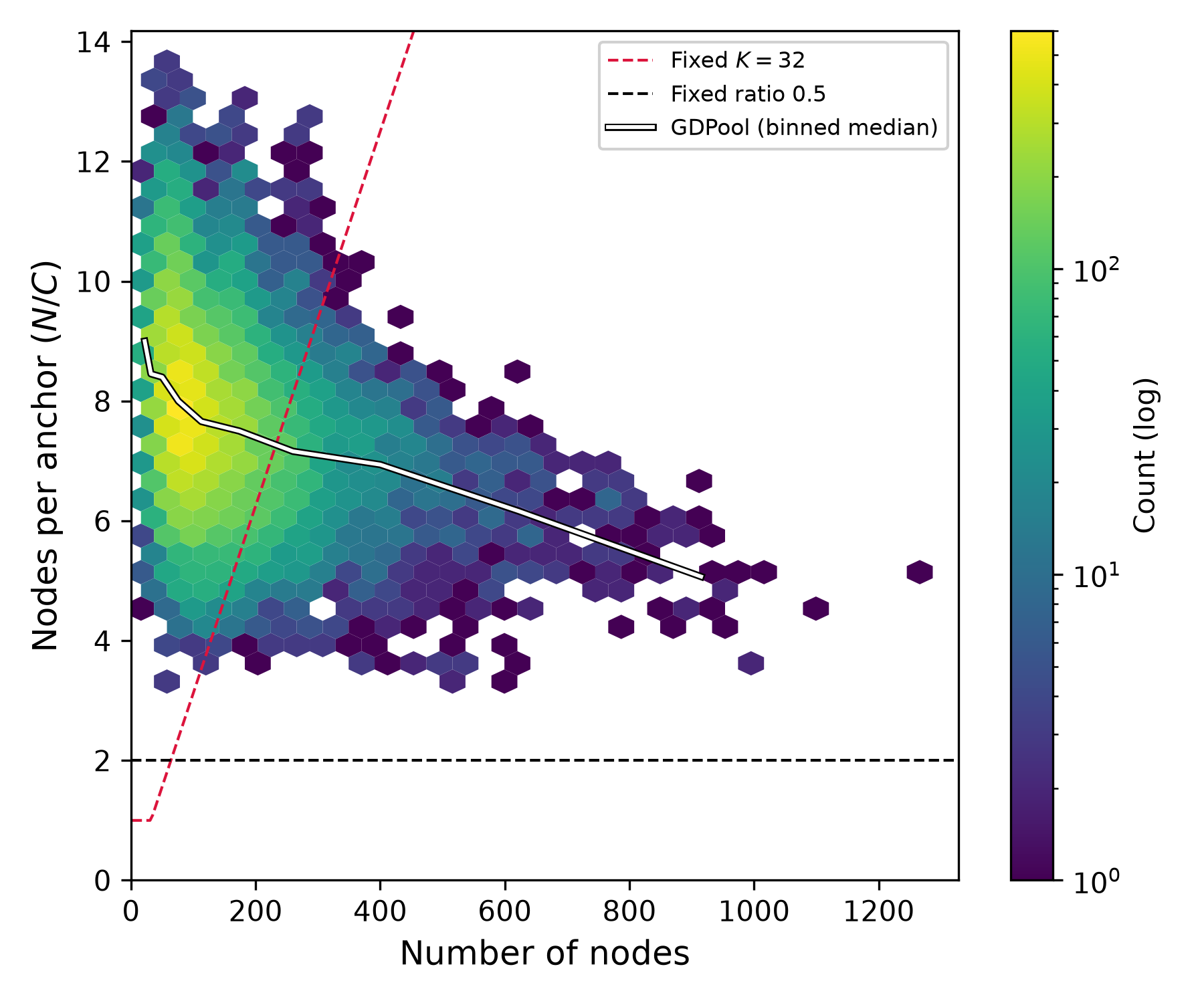}
        \caption{}
        \label{fig:node_anchor3}
    \end{subfigure}
    \caption{Anchor selection by GDPool at the first pooling level for CH$_4$@0.9 on the hMOF dataset. (a) Number of anchors. (b) Number of nodes per anchor. Red and black dashed lines indicate a fixed number of clusters and a fixed pooling ratio; color indicates graphs per bin on a logarithmic scale.}
    \label{fig:node_anchor_analysis}
\end{figure}

To test whether pooling granularity adapts to each graph, we examine the number of anchors $C$ selected by GDPool at the first pooling level as a function of the number of nodes $N$. As shown in Figure~\ref{fig:node_anchor1}, $C$ increases roughly in proportion to $N$, following neither a fixed number of clusters nor the default pooling ratio of 0.5 used by ratio-based methods. Around the trend, graphs of similar size receive different numbers of anchors, so the granularity reflects the structure of each graph. Figure~\ref{fig:node_anchor3} shows the resulting cluster size. The median number of nodes per anchor stays between 5 and 9, whereas a fixed $K=32$ would let it reach about 40. The number of anchors therefore scales with graph size to some extent, while graphs with the same number of nodes still receive different numbers of anchors depending on their structure. For further details on adaptive granularity, see Appendix~\ref{sec:appendix_granularity}. To verify that the variation does not stem from training randomness, we analyze models trained with three random seeds on the same graphs. The ICC of the number of selected anchors is 0.971, indicating that GDPool selects a consistent number of anchors for the same graph. As shown in Table~\ref{tab:jaccard}, the Jaccard similarity between anchor sets from different seeds is 0.581, far above the 0.071 obtained for random sets of the same size. The differences across graphs are therefore reproducible across seeds, reflecting the structure of each graph rather than the initialization (details in Appendix~\ref{sec:appendix_seed_robustness}).
We replace the graph-dependent anchor selection of GDPool by fixed top-$K$ and fixed-ratio rules while keeping the rest of the model unchanged. Table~\ref{tab:anchor_rule} in Appendix~\ref{sec:appendix_anchor_rule} shows that fixed rules at their default settings waste anchors or fail to coarsen the hierarchy; a ratio matched to the anchor counts of GDPool reaches comparable accuracy, yet still leaves 29\% of its first-level anchors unused, whereas GDPool activates 89\%. GDPool thus reaches this budget without tuning the number of clusters or the ratio, and its high activation rate is a prerequisite for the cluster-level analysis of the resulting hierarchy. We conduct a case study on HKUST-1~\citep{chui1999chemically}, a representative MOF for CO$_2$ adsorption. In its crystal structure, all 24 Cu paddlewheels and at least 90 of the 96 carboxylate-bearing ring carbons are represented by anchors in every training seed. These anchors coincide with the two kinds of vertices of the underlying net, the Cu paddlewheel nodes that host the primary CO$_2$ binding sites and the branch points of the linkers (Appendix~\ref{sec:appendix_case_study}).

\subsection{Gradient conflict diagnosis}
\label{sec:Gradient_Alignment_Analysis}

Section~\ref{sec:Gradient_Interactions_of_Hierarchical_Learning} analyzed how objectives at different pooling levels can produce optimization conflicts in hierarchical graph learning. We measure the gradient similarity of GDPool during training to analyze the effect of hierarchical supervision on gradient interactions. As shown in Figure~\ref{fig:gradient_similarity_base}, GDPool with hierarchical supervision maintains an alignment of approximately 0.6 to 0.8 throughout training, whereas conventional pooling methods with layer-wise supervision remain below 0.1 or become intermittently negative. We further compare the same GDPool architecture under hierarchical and layer-wise supervision. Figure~\ref{fig:gradient_similarity_gdpool} in Appendix~\ref{sec:appendix_gradient_alignment} shows the alignment over training epochs, where hierarchical supervision keeps the alignment both high and stable. Alignment depends not only on the supervision scheme but also on the anchor selection rule. Under the same hierarchical supervision, the default fixed rules lose gradient alignment and the matched ratio decays throughout training, whereas GDPool remains at about 0.6 in Figure~\ref{fig:gradient_similarity_as}. Hierarchical supervision and graph-dependent anchor selection thus mitigate the conflict from layer-wise supervision and ill-fitting budgets, respectively. The advantage of GDPool does not hinge on the supervision scheme: supervising only the final cumulative assignment matches hierarchical supervision in accuracy (MAE 0.152 vs.\ 0.151), whereas hierarchical supervision yields a markedly more compact and coherent hierarchy. Layer-wise supervision degrades both accuracy and structure (Appendix~\ref{sec:appendix_supervision}). Under identical hierarchical supervision, GDPool achieves lower MAE than DiffPool at either $\lambda$ on hMOF (Appendix~\ref{sec:appendix_hs}).

\subsection{Sensitivity analysis}
\label{sec:Sensitivity_to_Model_Components}

We analyze the sensitivity of GDPool to the entropy loss weight $\lambda$, the number of hierarchical pooling levels, and the GFFN expansion ratio. As shown in Table~\ref{tab:Effect_lambda}, MAE is insensitive to $\lambda$, whereas modularity increases overall as $\lambda$ grows. $K_{\mathrm{eff}}$ shows no clear trend but rises slightly overall. We adopt $\lambda=0.05$, which improves modularity without loss of predictive performance. To analyze the effect of pooling depth, we compare models trained with one and two pooling levels. As shown in Table~\ref{tab:Effect_depth}, GDPool achieves lower MAE than existing pooling methods in both settings, together with the best modularity at the first pooling level. The initial clustering behavior of GDPool therefore remains stable under changes in pooling depth. In Table~\ref{tab:ffn_ratio}, varying the GFFN expansion ratio makes little difference in performance. The analyses cover a GINEConv-based model and a SAGEConv-based model, with GDPool maintaining stable performance under hyperparameter and architectural changes in both cases.

\section{Conclusion}
\label{sec:Conclusion}

We proposed GDPool, which adaptively determines pooling granularity according to the structural characteristics of the input graph. GDPool selects a different number of anchors for each graph using the contextual importance computed by self-attention, constructing a coarse graph through soft-assignment of each node to the selected anchors. Hierarchical graph representations are thereby learned without fixing the number of clusters or the pooling ratio in advance. Our gradient alignment analysis reveals that auxiliary objectives at different pooling levels can send conflicting gradients to shared parameters. Supervising each level only through a local objective, as in existing practice, therefore does not guarantee coherent learning of the hierarchy.

On the four benchmarks, ARC-MOF, hMOF, ZINC, and QM9, GDPool consistently outperformed all compared pooling methods. In the layer-wise analysis of modularity and $K_{\mathrm{eff}}$, existing methods either retain the nominal number of clusters at every level or vary $K_{\mathrm{eff}}$ non-monotonically across levels, whereas $K_{\mathrm{eff}}$ of GDPool decreases monotonically with depth while maintaining high modularity. GDPool thus learns a granularity suited to the structure of the input graph rather than merely reducing the number of nodes.

Our study has several limitations. The number of anchors is determined by the mean attention score, so the selection criterion itself is not learned, and hierarchical supervision reuses existing clustering objectives rather than ones designed for cumulative assignments. Future work could explore learned granularity criteria and conflict-aware optimization, and evaluate GDPool on larger and more diverse graph-level tasks. Given the stable number of nodes per anchor across graph sizes and the high activation rates of the resulting clusters, GDPool can also be extended to assign predicted properties to cluster-level substructures.

\subsection*{AI use statement}

In this work, we have not used generative AI tools to generate experimental results, or make final scientific claims, and other required disclosure tasks are not applicable to this work. Additionally, we used generative AI tools for language editing and proofreading. We have reviewed all AI-assisted work. We take responsibility for the final content of this work, including text, claims or artifacts produced with the aid of generative AI.

\subsection*{Reproducibility statement}

The detailed pseudocode of GDPool is provided in Appendix~\ref{sec:appendix_gdpool_algorithm}. Details necessary to reproduce our experiments, including dataset descriptions, model configurations, training settings, and evaluation metrics, are provided in Section~\ref{sec:Experiments}, Appendix~\ref{sec:appendix_implementation_details}, Appendix~\ref{sec:appendix_datasets_details} and Appendix~\ref{sec:appendix_metrics}. Additional experimental results, including gradient alignment and sensitivity analyses, are reported in the Appendix. An anonymized implementation and configuration files are provided as supplementary materials.


\bibliography{iclr2027_conference}
\bibliographystyle{iclr2027_conference}

\appendix

\section{GDPool algorithm}
\label{sec:appendix_gdpool_algorithm}

GDPool transforms $G=(X,A)$ into $G'=(X',A')$ based on $C$ anchors derived from the structural characteristics of the input graph. Since $C$ is determined dynamically for each graph, GDPool can construct hierarchical graph representations with different pooling granularities for different input graphs. The overall procedure of the GDPool is summarized in the pseudocode in Algorithm~\ref{alg:gdpool}.
\begin{algorithm}[ht]
\caption{Graph-dependent anchor selection and assignment}
\label{alg:gdpool}
\begin{algorithmic}[1]
\Require node features $X \in \mathbb{R}^{N \times D}$, valid-node mask $M \in \{0,1\}^{N}$
\Ensure assignment matrix $S \in \mathbb{R}^{N \times C}$, anchor set $\mathcal{A}=\{a_1,\ldots,a_C\}$
\State $\bar{X} \gets \operatorname{LayerNorm}(X)$
\State $Q \gets \bar{X}, \quad K \gets \bar{X}$
\State $R \gets QK^{\top}/\sqrt{D}$
\State $R_{ij} \gets -\infty$ if $M_iM_j=0$
\State $H \gets \operatorname{Softmax}_{j}(R)$
\State $p_j \gets \sum_i H_{ij}$
    \Comment{Node importance}
\State $N_v \gets \sum_j M_j$
\State $\bar{p} \gets \frac{1}{N_v} \sum_{j:M_j=1} p_j$
    \Comment{Mean node importance}
\State $\mathcal{A} \gets \{j \mid p_j > \bar{p},\; M_j=1\}$
\State $a^* \gets \arg\max_{j:M_j=1} p_j$
\State $\mathcal{A} \gets \mathcal{A} \cup \{a^*\}$
    \Comment{Ensure at least one anchor}
\State $R^{\mathcal{A}}_{ik} \gets R_{a_k i}, \quad a_k \in \mathcal{A}$
\State $S_{ik} \gets \operatorname{Softmax}_{k} \left(R^{\mathcal{A}}_{ik}\right)$
    \Comment{Graph-dependent soft assignment}
\State $S_{ik} \gets 0$ if $M_i=0$
\State \Return $S, \mathcal{A}$
\end{algorithmic}
\end{algorithm}

\section{Implementation details}
\label{sec:appendix_implementation_details}

\subsection{GDPool Layer}
\label{sec:appendix_gdpool_layer}

\begin{figure}[ht]
\centering
\includegraphics[width=0.8\linewidth]{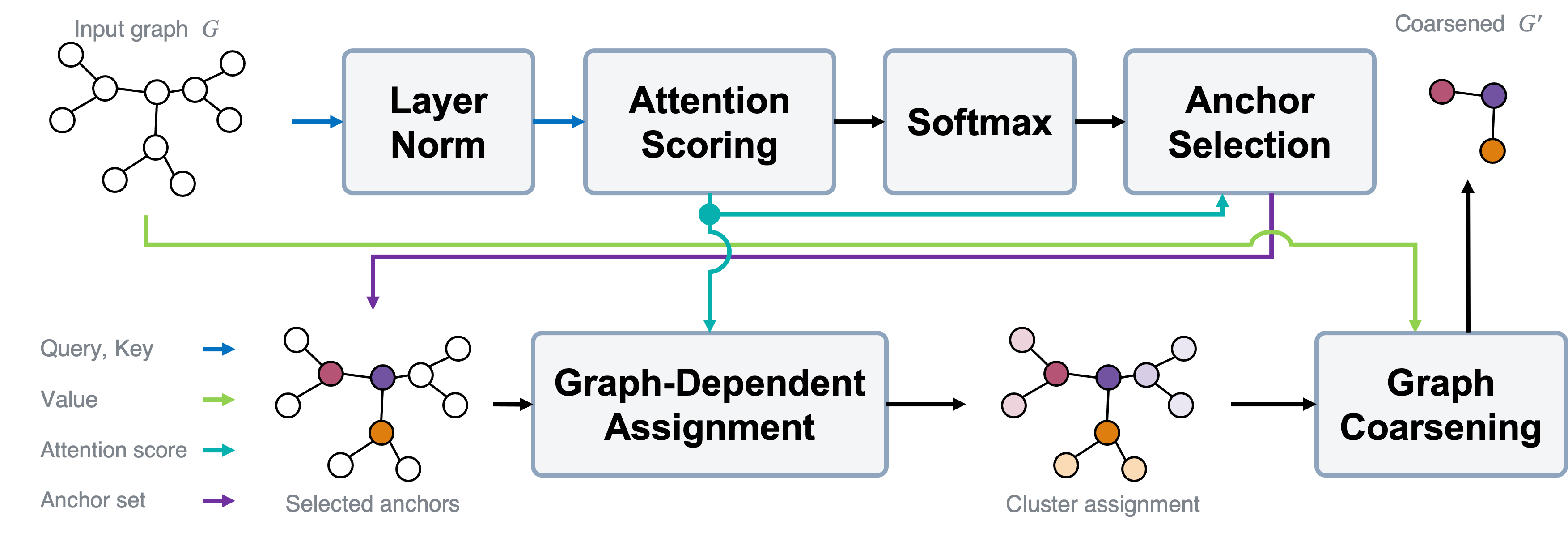}
\caption{Scheme of the GDPool layer.}
\label{fig:gdpool_layer}
\end{figure}

A schematic depiction of the GDPool layer is illustrated in Figure~\ref{fig:gdpool_layer}. GDPool first uses the normalized node representations as queries and keys to score node importance, and selects anchor nodes for each graph based on these scores. It then derives a soft assignment of each node to the anchors and aggregates the values of the original nodes according to this assignment to produce the coarsened graph.

\subsection{Model configurations}
\label{sec:appendix_model_configurations}

The embedding dimension of the node representations was set to 64, and the batch size was fixed at 64. All models were trained with the AdamW optimizer~\citep{loshchilov2017decoupled} with a weight decay of $10^{-4}$. The learning rate followed a cosine annealing schedule~\citep{loshchilov2016sgdr}, with the maximum and minimum learning rates set to $10^{-3}$ and $10^{-5}$, respectively. On the ARC-MOF and hMOF datasets, all models were trained for 100 epochs with three hierarchical pooling levels; for pooling methods that require the number of clusters to be specified in advance, the number of clusters at each level was set to 32, 16, and 8, respectively. On the ZINC and QM9 datasets, all models were trained for 300 epochs with two hierarchical pooling levels, and the number of clusters at each level was set to 12 and 6, respectively, for such methods. For the single-level models in Table~\ref{tab:Effect_depth}, the number of clusters was set to 12. For pooling-ratio-based methods, the default pooling ratio of each method was used on all datasets. All experiments were repeated three times with different random seeds, and the same three seeds were used for every method. For each run, the checkpoint with the lowest validation MAE is used for evaluation on the test set. Finally, all experiments were conducted on a single NVIDIA RTX PRO 6000 GPU.

\section{Complexity and computational cost}
\label{sec:appendix_complexity}

Table~\ref{tab:complexity} reports the training time per epoch and the peak GPU memory measured on the hMOF dataset. In terms of training time, GDPool takes $236.21$ seconds per epoch, which is faster than MemPool, comparable to MaxCutPool, and about $1.8\times$ that of SAGPool, the fastest method. It is also more than $8\times$ faster than EdgePool, which performs iterative edge contraction. In contrast, its peak memory usage of $10.56$\,GB is higher than that of all other methods. This is because GDPool must retain both the attention logits and the softmax outputs, and additionally constructs the assignment to the anchors on top of them. Moreover, since the graphs in a mini-batch are padded to $C_{\max} = \max_b C_b$, where $C_b$ is the number of anchors in the $b$-th graph, a single graph with many anchors forces the memory of the entire batch to be allocated accordingly. As shown in Figure~\ref{fig:node_anchor_analysis}, the number of anchors scales with graph size and the size distribution of hMOF graphs is right-skewed. The distribution of anchor counts has a long right tail, so the actual memory usage is governed by the maximum within a batch rather than by the average number of anchors. The relatively large standard deviation of the training time is likewise attributable to the variation of $C_{\max}$ across batches. Since the number of anchors is determined during the forward pass, batches cannot be formed by exact anchor count at the data loading stage. Because the number of anchors scales with graph size, however, bucketing graphs by size would largely reduce this overhead; we leave this to future work.
\begin{table}[ht]
\caption{Memory complexity and measured cost on the hMOF dataset. $N$ and $E$ denote the number of nodes and edges. Time is the duration of one training epoch. Peak memory is the maximum GPU memory observed during that epoch.}
\label{tab:complexity}
\centering
\begin{tabular}{lccc}
\toprule
Method & Memory & Time (sec) & Peak mem. (GB) \\
\midrule
DiffPool     & $O(N^2)$ & $160.72 {\scriptstyle \pm 2.43}$  & 3.90 \\
MinCutPool   & $O(N^2)$ & $188.56 {\scriptstyle \pm 3.16}$  & 3.73 \\
MemPool      & $O(N^2)$ & $293.36 {\scriptstyle \pm 2.83}$  & 3.68 \\
BN-Pool	     & $O(N^2)$    & $153.07 {\scriptstyle \pm 0.87}$  & 4.35 \\
TopKPool     & $O(N + E)$    & $147.15 {\scriptstyle \pm 1.03}$  & 2.89 \\
SAGPool      & $O(N + E)$    & $128.98 {\scriptstyle \pm 0.77}$  & 2.80 \\
ASAPool      & $O(N + E)$    & $162.76 {\scriptstyle \pm 1.41}$  & 4.01 \\
MaxCutPool   & $O(N + E)$    & $213.71 {\scriptstyle \pm 15.55}$  & 2.98 \\
EdgePool     & $O(N + E)$    & $1939.43 {\scriptstyle \pm 332.43}$  & 2.82 \\
\midrule
\textbf{GDPool}       & $O(N^2)$ & $236.21 {\scriptstyle \pm 21.80}$ & 10.56 \\
\bottomrule
\end{tabular}
\end{table}

\section{Datasets details}
\label{sec:appendix_datasets_details}

The statistics of the benchmark datasets used in this paper, ARC-MOF, hMOF, ZINC, and QM9, are reported in Table~\ref{tab:dataset_statistics}. For ARC-MOF and hMOF, the labels are gas uptake capacities obtained from GCMC simulations at the pressures indicated in each column header, reported in mol/kg. The hMOF isotherms were obtained from MOFX-DB~\citep{bobbitt2023mofx}, which hosts simulated isotherms for the hypothetical MOF structures of \citet{wilmer2012large}. For ZINC we use the 12{,}000 graph subset, whose label is the constrained solubility $\mathrm{logP}-\mathrm{SAS}-\mathrm{cycles}$. For QM9, we predict the dipole moment $\mu$, reported in Debye. We use all hMOF structures with CO$_2$ adsorption data in MOFX-DB (137,652 structures) and, for ARC-MOF, the ARCMOF\_20241004 release with all structures that have post-combustion CO$_2$ adsorption data (post\_comb\_vsa-CO2.csv; 242,390 structures), without further filtering. The hMOF dataset is randomly split into training, validation, and test sets with a 70\%/15\%/15\% ratio. The ARC-MOF dataset is split with the same ratio by a coverage-aware random procedure: for each element, from rarest to most common, one unassigned structure containing it is placed in each split so that every element appears in all three splits, and the remaining structures are assigned at random. QM9 is randomly split with an 80\%/10\%/10\% ratio, and for ZINC we use the default split provided by PyTorch Geometric. The same splits are used for all compared methods. ARC-MOF contains several variants of the same parent structure (e.g., different solvent-removal treatments); because a single split is shared by all methods, the methods are compared under identical conditions, but absolute errors may be optimistic. The 3D atomic coordinates available in QM9 are not used. All targets are used in their native units, without further normalization or rescaling. Reported errors are therefore directly interpretable in physical units, but are not comparable to task-specialized models that are tuned per dataset and employ target normalization.
\begin{table}[ht]
\caption{Statistics of datasets.}
\label{tab:dataset_statistics}
\begin{center}
\begin{tabular}{lccc}
\toprule
\textbf{Dataset} & \textbf{\# Graphs} & \textbf{Avg. \# Nodes} & \textbf{Avg. \# Edges} \\
\midrule
ARC-MOF  & 242,390 & 157.57 & 1890.8 \\
hMOF  & 137,652 & 141.39 & 1696.7 \\
ZINC  & 12,000 & 23.2 & 49.8 \\
QM9   & 130,831 & 18.0 & 37.3 \\
\bottomrule
\end{tabular}
\end{center}
\end{table}

\section{Additional results for property prediction}
\label{sec:appendix_additional_property_prediction}

The main paper reports MAE results for property prediction, while RMSE and $R^2$ are also used as evaluation metrics. The corresponding RMSE and $R^2$ results are provided in Tables~\ref{tab:performance_comparison_rmse} and \ref{tab:performance_comparison_r2}, respectively, for completeness. ASAPool and MaxCutPool did not train stably in our setting. ASAPool converges to a near-constant predictor on both hMOF targets, yielding negative $R^2$ in Table~\ref{tab:performance_comparison_r2}, and MaxCutPool shows the largest variance among all compared methods. Both were given the same backbone, training budget, and hyperparameter search as every other baseline, and we report their results as obtained rather than excluding them. Two factors may contribute. First, their node-selection mechanisms may be mismatched to the large, dense material graphs used here. Second, both are typically evaluated with a single pooling layer, whereas our setting stacks two or three; consistently, Table~\ref{tab:Effect_depth} shows that all four node-selection baselines degrade as pooling depth increases from one to two levels, with MaxCutPool degrading the most, whereas GDPool improves. These observations concern the behavior of these methods under deep hierarchical stacking and should not be read as a general assessment of them.
\begin{table}[ht]
\caption{RMSE of graph pooling methods on the benchmark datasets.}
\label{tab:performance_comparison_rmse}
\begin{center}
\begin{tabular}{lccccc}
\toprule
\multirow{2}{*}{\textbf{Method}} & \multicolumn{1}{c}{\textbf{ARC-MOF}} & \multicolumn{2}{c}{\textbf{hMOF}} & \multicolumn{1}{c}{\textbf{ZINC}} & \multicolumn{1}{c}{\textbf{QM9}} \\
\cmidrule(lr){2-2}\cmidrule(lr){3-4}\cmidrule(lr){5-5}\cmidrule(lr){6-6}
& \multicolumn{1}{c}{CO$_2$@0.15} & \multicolumn{1}{c}{CO$_2$@0.1} & \multicolumn{1}{c}{CH$_4$@0.9} & \multicolumn{1}{c}{Solubility} & \multicolumn{1}{c}{Dipole Moment} \\
\midrule
DiffPool
& $0.356 {\scriptstyle \pm 0.031}$ & $0.316 {\scriptstyle \pm 0.002}$ 
& $0.247 {\scriptstyle \pm 0.019}$ & $0.897 {\scriptstyle \pm 0.004}$ 
& $0.696 {\scriptstyle \pm 0.013}$ \\
MinCutPool
& $0.295 {\scriptstyle \pm 0.003}$ & $0.262 {\scriptstyle \pm 0.003}$ 
& $0.200 {\scriptstyle \pm 0.003}$ & $0.811 {\scriptstyle \pm 0.010}$ 
& $0.659 {\scriptstyle \pm 0.004}$ \\
MemPool
& $0.293 {\scriptstyle \pm 0.001}$ & $0.262 {\scriptstyle \pm 0.001}$ 
& $0.202 {\scriptstyle \pm 0.001}$ & $0.826 {\scriptstyle \pm 0.004}$ 
& $0.654 {\scriptstyle \pm 0.004}$ \\
BN-Pool
& $0.297 {\scriptstyle \pm 0.011}$ & $0.263 {\scriptstyle \pm 0.002}$ 
& $0.205 {\scriptstyle \pm 0.002}$ & $0.824 {\scriptstyle \pm 0.011}$ 
& $0.665 {\scriptstyle \pm 0.003}$ \\
TopKPool
& $0.325 {\scriptstyle \pm 0.005}$ & $0.275 {\scriptstyle \pm 0.003}$ 
& $0.213 {\scriptstyle \pm 0.002}$ & $0.863 {\scriptstyle \pm 0.040}$ 
& $0.682 {\scriptstyle \pm 0.016}$ \\
SAGPool
& $0.309 {\scriptstyle \pm 0.002}$ & $0.273 {\scriptstyle \pm 0.001}$ 
& $0.216 {\scriptstyle \pm 0.004}$ & $0.941 {\scriptstyle \pm 0.029}$ 
& $0.698 {\scriptstyle \pm 0.008}$ \\
ASAPool
& $0.670 {\scriptstyle \pm 0.621}$ & $0.733 {\scriptstyle \pm 0.000}$ 
& $0.732 {\scriptstyle \pm 0.000}$ & $1.003 {\scriptstyle \pm 0.072}$ 
& $0.709 {\scriptstyle \pm 0.002}$ \\
MaxCutPool
& $0.647 {\scriptstyle \pm 0.273}$ & $0.597 {\scriptstyle \pm 0.037}$ 
& $0.445 {\scriptstyle \pm 0.182}$ & $1.136 {\scriptstyle \pm 0.090}$ 
& $0.959 {\scriptstyle \pm 0.142}$ \\
EdgePool
& $0.304 {\scriptstyle \pm 0.001}$ & $0.264 {\scriptstyle \pm 0.002}$ 
& $0.207 {\scriptstyle \pm 0.003}$ & $0.813 {\scriptstyle \pm 0.005}$ 
& $0.661 {\scriptstyle \pm 0.002}$ \\
\midrule
\textbf{GDPool}
& $\textbf{0.282}{\scriptstyle \pm 0.003}$ & $\textbf{0.258}{\scriptstyle \pm 0.004}$
& $\textbf{0.198}{\scriptstyle \pm 0.001}$ & $\textbf{0.580}{\scriptstyle \pm 0.039}$
& $\textbf{0.601}{\scriptstyle \pm 0.004}$ \\
\bottomrule
\end{tabular}
\end{center}
\end{table}

\begin{table}[ht]
\caption{$R^2$. of graph pooling methods on the benchmark datasets.}
\label{tab:performance_comparison_r2}
\begin{center}
\begin{tabular}{lccccc}
\toprule
\multirow{2}{*}{\textbf{Method}} & \multicolumn{1}{c}{\textbf{ARC-MOF}} & \multicolumn{2}{c}{\textbf{hMOF}} & \multicolumn{1}{c}{\textbf{ZINC}} & \multicolumn{1}{c}{\textbf{QM9}} \\
\cmidrule(lr){2-2}\cmidrule(lr){3-4}\cmidrule(lr){5-5}\cmidrule(lr){6-6}
& \multicolumn{1}{c}{CO$_2$@0.15} & \multicolumn{1}{c}{CO$_2$@0.1} & \multicolumn{1}{c}{CH$_4$@0.9} & \multicolumn{1}{c}{Solubility} & \multicolumn{1}{c}{Dipole Moment} \\
\midrule
DiffPool
& $0.894 {\scriptstyle \pm 0.018}$ & $0.786 {\scriptstyle \pm 0.003}$ 
& $0.855 {\scriptstyle \pm 0.022}$ & $0.802 {\scriptstyle \pm 0.002}$ 
& $0.786 {\scriptstyle \pm 0.008}$ \\
MinCutPool
& $0.928 {\scriptstyle \pm 0.002}$ & $0.853 {\scriptstyle \pm 0.004}$ 
& $0.905 {\scriptstyle \pm 0.003}$ & $0.838 {\scriptstyle \pm 0.004}$ 
& $0.808 {\scriptstyle \pm 0.003}$ \\
MemPool
& $0.929 {\scriptstyle \pm 0.001}$ & $0.853 {\scriptstyle \pm 0.002}$ 
& $0.903 {\scriptstyle \pm 0.001}$ & $0.832 {\scriptstyle \pm 0.002}$ 
& $0.811 {\scriptstyle \pm 0.002}$ \\
BN-Pool
& $0.927 {\scriptstyle \pm 0.006}$ & $0.852 {\scriptstyle \pm 0.003}$ 
& $0.900 {\scriptstyle \pm 0.002}$ & $0.833 {\scriptstyle \pm 0.005}$ 
& $0.805 {\scriptstyle \pm 0.002}$ \\
TopKPool
& $0.912 {\scriptstyle \pm 0.003}$ & $0.838 {\scriptstyle \pm 0.003}$ 
& $0.893 {\scriptstyle \pm 0.002}$ & $0.817 {\scriptstyle \pm 0.017}$ 
& $0.794 {\scriptstyle \pm 0.010}$ \\
SAGPool
& $0.921 {\scriptstyle \pm 0.001}$ & $0.840 {\scriptstyle \pm 0.001}$ 
& $0.890 {\scriptstyle \pm 0.004}$ & $0.782 {\scriptstyle \pm 0.013}$ 
& $0.784 {\scriptstyle \pm 0.005}$ \\
ASAPool
& $0.414 {\scriptstyle \pm 0.875}$ & $-0.150 {\scriptstyle \pm 0.000}$ 
& $-0.267 {\scriptstyle \pm 0.000}$ & $0.752 {\scriptstyle \pm 0.035}$ 
& $0.778 {\scriptstyle \pm 0.002}$ \\
MaxCutPool
& $0.612 {\scriptstyle \pm 0.263}$ & $0.235 {\scriptstyle \pm 0.096}$ 
& $0.478 {\scriptstyle \pm 0.338}$ & $0.681 {\scriptstyle \pm 0.049}$ 
& $0.587 {\scriptstyle \pm 0.117}$ \\
EdgePool
& $0.924 {\scriptstyle \pm 0.000}$ & $0.851 {\scriptstyle \pm 0.002}$ 
& $0.899 {\scriptstyle \pm 0.002}$ & $0.838 {\scriptstyle \pm 0.002}$ 
& $0.807 {\scriptstyle \pm 0.001}$ \\
\midrule
\textbf{GDPool}
& $\textbf{0.934}{\scriptstyle \pm 0.001}$ & $\textbf{0.857}{\scriptstyle \pm 0.004}$
& $\textbf{0.907}{\scriptstyle \pm 0.001}$ & $\textbf{0.917}{\scriptstyle \pm 0.011}$
& $\textbf{0.840}{\scriptstyle \pm 0.002}$ \\
\bottomrule
\end{tabular}
\end{center}
\end{table}

\section{Evaluation metrics of hierarchical clustering behavior}
\label{sec:appendix_metrics}

Modularity is computed directly from the soft assignment as
\begin{equation}
    Q = \frac{1}{2m}\,\mathrm{tr}\!\left(S^{\top} B S\right), \qquad B = A - \frac{dd^{\top}}{2m},
\end{equation}
where $B$ is the modularity matrix~\citep{newman2006modularity}, $d$ is the degree vector, and $m$ is the number of edges. For hard assignments this reduces to the standard modularity, and near-uniform soft assignments yield $Q \approx 0$. At level $l>1$, $A$ denotes the coarse adjacency produced by the preceding level, so modularity values at deeper levels are computed on progressively smaller graphs and are not directly comparable across methods that coarsen at different rates. 

The effective number of clusters $K_\mathrm{eff}$ measures how many clusters are actually used~\citep{zheng2023online}. For each graph, the total assignment mass received by each cluster is normalized into a distribution over clusters, and $K_\mathrm{eff}$ is the perplexity of this distribution, i.e., the exponential of its entropy. It equals the number of clusters when all clusters receive equal mass and approaches one when the assignment collapses onto a single cluster. Both metrics are computed per graph and averaged over the test set.

\section{Anchor selection analyses}
\label{sec:appendix_anchor_selection_analyses}

\subsection{Adaptive pooling granularity}
\label{sec:appendix_granularity}

We discuss in detail the visualization of pooling granularity presented in Section~\ref{sec:adaptive_anchor_selection}. For each graph in the hMOF test set, we recorded the number of original nodes $N$ together with the number of anchors $C$ selected at the first pooling level, averaged over three seeds. Note that, unlike $K_\mathrm{eff}$, $C$ refers to the number of anchors actually selected by the anchor selection step. As shown in Figure~\ref{fig:node_anchor1}, $C$ grows approximately linearly with $N$. A linear fit through the origin gives $C \approx 0.15\,N$ (Pearson $r = 0.96$), and $C/N$ lies between 0.10 and 0.17 for 80\% of the graphs (10th--90th percentiles, median 0.13). GDPool thus behaves like a ratio-based pooling method whose ratio is not set in advance but emerges from the anchor criterion $p_j > \bar p$, and the emergent ratio differs from both a fixed number of clusters ($K=32$) and the default pooling ratio of 0.5. Around this trend, graphs of similar size receive different numbers of anchors. Within a 50-node size range, $C$ varies by a factor of about 1.6 between the 10th and 90th percentiles, so the granularity reflects not only the size of each graph but also its structure. Figure~\ref{fig:node_anchor3} shows the same data as nodes per anchor, $N/C$. GDPool keeps this quantity within a narrow range, whereas a fixed $K=32$ would let it grow from 1 to 40 across the graph sizes in the test set. On average, GDPool selects 19.60 anchors per graph, of which $K_\mathrm{eff} = 17.35$ are active, corresponding to an activation rate of approximately 89\%. In contrast, existing pooling methods either fail to form proper clusters or activate fewer than 20\% of their clusters. This shows that GDPool not only allocates anchors in proportion to graph size but also actually utilizes most of the clusters it forms.

The motivation for graph-dependent granularity is interpretability. Properties of both materials and molecules are governed by local substructures, such as metal nodes and linkers in MOFs or functional groups in molecules, and we ultimately aim to attribute predicted properties to such substructures at the cluster level, which we leave to future work. This requires cluster sizes set by local structure rather than by graph size. A fixed number of clusters cannot provide this. A fixed pooling ratio keeps the cluster size constant but assigns the same budget to every graph of the same size regardless of its structure, whereas GDPool varies the budget among graphs of the same size.

\subsection{Seed robustness of anchor selection}
\label{sec:appendix_seed_robustness}

To assess the robustness of anchor selection to the choice of random seed, we measured the consistency of both the number of selected anchors and their node identities across three different seeds. To quantify the consistency of the number of selected anchors, we used the one-way random-effects ICC, i.e., ICC(1,1) in the notation of \citet{shrout1979intraclass}. Treating each graph as the unit of measurement and each random seed as a repeated measurement, we compute the ICC as follows:
\begin{equation}
    \mathrm{ICC} = \frac{\mathrm{MS}_B-\mathrm{MS}_W}{\mathrm{MS}_B+(s-1)\mathrm{MS}_W},
    \label{eq:icc}
\end{equation}
where $\mathrm{MS}_{B}$ is the between-graph mean square, which captures the variation across graphs, $\mathrm{MS}_{W}$ is the within-graph mean square, which captures the variation across seeds within the same graph, and $s$ is the number of random seeds. An ICC close to 1 therefore indicates that the variation arising from differences between graphs dominates the variation induced by the random seed, i.e., that the number of anchors is stable with respect to the seed. This experiment was conducted on the first pooling layer of the model trained on the CH$_4$@0.9 of the hMOF dataset. The within-graph standard deviation across seeds ($\sqrt{\mathrm{MS}_{W}}$) was $3.02$, whereas the standard deviation of the per-graph mean anchor count was $17.54$, yielding an ICC of $0.971$. This demonstrates that the number of anchors selected for the same graph remains consistent even when the random seed is changed. Because $C$ scales with $N$, part of this agreement reflects graph size. The ICC of the size-normalized count $C/N$ is 0.67, so the graph-specific deviations from the size trend are also largely reproducible across seeds.

To assess the consistency of the selected nodes as well as their number, we additionally measured the Jaccard similarity between anchor sets. For each pair of seeds, we computed the Jaccard similarity per graph and averaged it over all graphs; the final score is the mean over the three seed pairs. Table~\ref{tab:jaccard} reports this score together with a random baseline that selects the same number of nodes uniformly at random for each graph. The anchors selected under different seeds overlap far more than expected from random selection. These results show that the anchor selection of GDPool is stable under seed changes in terms of both the number of anchors and the identity of the selected nodes.
\begin{table}[ht]
\caption{Jaccard similarity of anchor sets across different random seeds and the corresponding random-selection baselines for CH$_4$@0.9 on the hMOF dataset.}
\label{tab:jaccard}
\begin{center}
\begin{tabular}{lcccc}
\toprule
\textbf{Method} & \textbf{Pair 1} & \textbf{Pair 2} & \textbf{Pair 3} & \textbf{Avg.} \\
\midrule
Random  & 0.071 & 0.074 & 0.069 & 0.071 \\
GDPool & \textbf{0.554} & \textbf{0.586} & \textbf{0.602} & \textbf{0.581} \\
\bottomrule
\end{tabular}
\end{center}
\end{table}

\subsection{Effect of the anchor selection rule}
\label{sec:appendix_anchor_rule}

Table~\ref{tab:anchor_rule} replaces the adaptive anchor rule by fixed rules while keeping the rest of the model unchanged. At their default settings, both fixed rules fail to form a coherent hierarchy. With $K=32/16/8$, only 14.7 of 32 first-level anchors are active and $K_\mathrm{eff}$ does not decrease from the first to the second level (14.7 to 14.8), so the first level barely coarsens the graph. With a ratio of 0.5, 70 anchors are allocated at the first level but 17 are active, $K_\mathrm{eff}$ increases from the first to the second level (17.5 to 32.3), and accuracy degrades (MAE 0.131, $p=0.02$). A ratio matched to the anchor counts of GDPool (0.15 at the first level and 0.5 thereafter) reaches comparable accuracy (MAE 0.124 vs.\ 0.125, within seed variance) and a monotonically decreasing $K_\mathrm{eff}$, but activates only 71\% of its first-level anchors (15.4 of 21.5), because the same fraction is applied to every graph of a given size regardless of its structure. This setting was moreover read off the output of GDPool and would otherwise require a per-level, per-dataset search. GDPool obtains the same budget without any setting, allocates it per graph, and activates 89\% of its first-level anchors (17.4 of 19.7). Using the resulting clusters in downstream analysis requires both properties. Modularity measures whether clusters follow the graph structure, and the activation rate measures whether allocated clusters are actually populated. A cluster that follows the structure but is empty cannot be analyzed, and a cluster that is populated but does not follow the structure cannot be interpreted. Since the anchor-based rules have similar modularity (0.38--0.43 at the first two levels), the activation rate is what separates them for downstream use, and GDPool is highest in this respect.
\begin{table}[ht]
\centering
\caption{Effect of the anchor selection rule on GDPool for CO$_2$@0.1 on the hMOF dataset. Default settings are $K=32/16/8$ and a ratio of 0.5 at every level; the matched ratio uses 0.15 at the first level and 0.5 thereafter, which reproduces the mean number of anchors of GDPool. Active is the fraction of allocated anchors that are used.  \#Anchors, $K_\mathrm{eff}$, and Active are reported per pooling level (L1\,/\,L2\,/\,L3).}
\label{tab:anchor_rule}
\small
\setlength{\tabcolsep}{4pt}
\begin{tabular}{lcccc}
\toprule
\textbf{Anchor rule} & \textbf{MAE} & \textbf{\#Anchors} & $\boldsymbol{K}_{\mathrm{eff}}$ & \textbf{Active (\%)} \\
\midrule
Fixed $K$ & 0.126$\pm$0.002 & 31.9 / 16.0 / 8.0 & 14.7 / 14.8 / 7.5 & 46.1 / 92.5 / 94.2 \\
Fixed ratio & 0.131$\pm$0.002 & 70.4 / 35.4 / 18.0 & 17.5 / 32.3 / 16.2 & 24.8 / 91.1 / 90.0 \\
Fixed ratio (matched) & \textbf{0.124}$\pm$0.001 & 21.5 / 11.0 / 5.8 & 15.4 / 10.4 / 5.6 & 71.3 / 94.7 / 97.5 \\
\textbf{GDPool} & 0.125$\pm$0.002 & 19.7 / 10.0 / 5.7 & 17.4 / 9.6 / 5.5 & \textbf{88.6} / \textbf{95.5} / \textbf{97.8} \\
\bottomrule
\end{tabular}
\end{table}

\subsection{Case study on the HKUST-1}
\label{sec:appendix_case_study}

To examine where the anchor criterion places anchors in a real framework, we apply the GDPool model trained on hMOF (CO$_2$ at 0.1\,bar) to the experimental crystal structure of HKUST-1, Cu$_3$(BTC)$_2$~\citep{chui1999chemically}. Coordinated and free water molecules are removed with the CoRE MOF cleaning procedure~\citep{chung2019advances}, which yields the activated framework with 624 atoms (Cu$_{48}$C$_{288}$O$_{192}$H$_{96}$); the structure is then converted into a graph with the same preprocessing as the hMOF dataset. Table~\ref{tab:hkust_seeds} summarizes the first-level anchors for the three training seeds. For seed 42, the first pooling level selects 148 anchors: all 48 Cu atoms, 95 of the 96 carboxylate-bearing ring carbons, one symmetry class of the BTC linker, and five further atoms (one C--H ring carbon, three O, one H), so essentially every tritopic linker contributes three anchors at the positions where its arms leave the ring. The pattern holds across seeds: the 142--148 anchors are drawn from the same two symmetry classes, namely 40--48 Cu atoms, with every one of the 24 Cu paddlewheels retaining at least one Cu anchor, and 90--95 of the 96 carboxylate-bearing carbons, whereas none of the 96 carboxylate carbons is selected in any seed. The anchors therefore coincide with the two kinds of vertices of the underlying net, the Cu paddlewheel nodes and the branch points of the linkers, and the first-level clusters partition the framework into metal coordination spheres and linker arms. 
\begin{table}[ht]
\centering
\caption{Anchors of GDPool on HKUST-1 for the three training seeds of the hMOF CO$_2$@0.1 model. C$_{\mathrm{ipso}}$: ring carbons bearing a carboxylate; C$_{\mathrm{H}}$: ring carbons bearing hydrogen; C$_{\mathrm{COO}}$: carboxylate carbons. Paddlewheels: Cu$_2$ paddlewheels with at least one Cu anchor.}
\label{tab:hkust_seeds}
\begin{tabular}{lcccccccc}
\toprule
\textbf{Seed} & \textbf{\#Anchors} & \textbf{Cu} & \textbf{Paddlewheels} & \textbf{C}$_{\mathrm{ipso}}$ & \textbf{C}$_{\mathrm{H}}$ & \textbf{C}$_{\mathrm{COO}}$ & \textbf{O} & \textbf{H} \\
\midrule
42 & 148 (23.7\%) & 48 & 24 & 95 & 1 & 0 & 3 & 1 \\
43 & 142 (22.8\%) & 40 & 24 & 90 & 7 & 0 & 4 & 1 \\
44 & 144 (23.1\%) & 47 & 24 & 94 & 0 & 0 & 3 & 0 \\
\bottomrule
\end{tabular}
\end{table}

\begin{figure}[ht]
    \centering
    \begin{subfigure}[t]{0.43\linewidth}
        \centering
        \includegraphics[width=\linewidth]{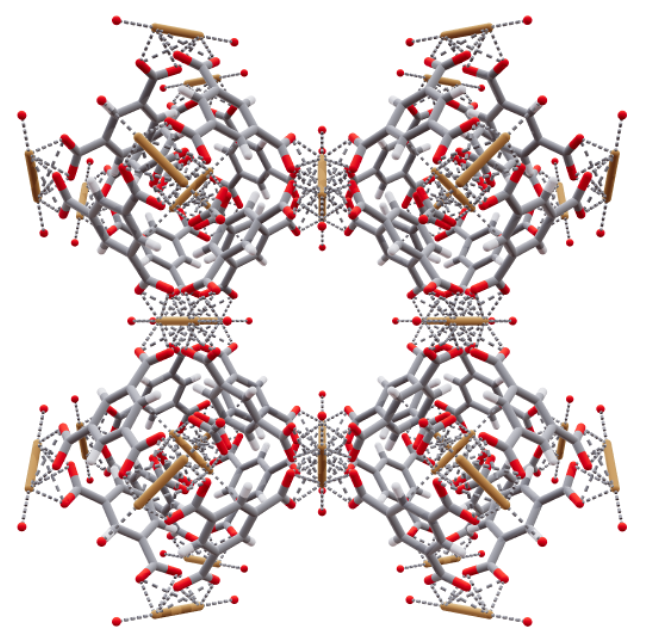}
        \caption{}
        \label{fig:hkust_base}
    \end{subfigure}%
    \hspace{0.05\linewidth}%
    \begin{subfigure}[t]{0.43\linewidth}
        \centering
        \includegraphics[width=\linewidth]{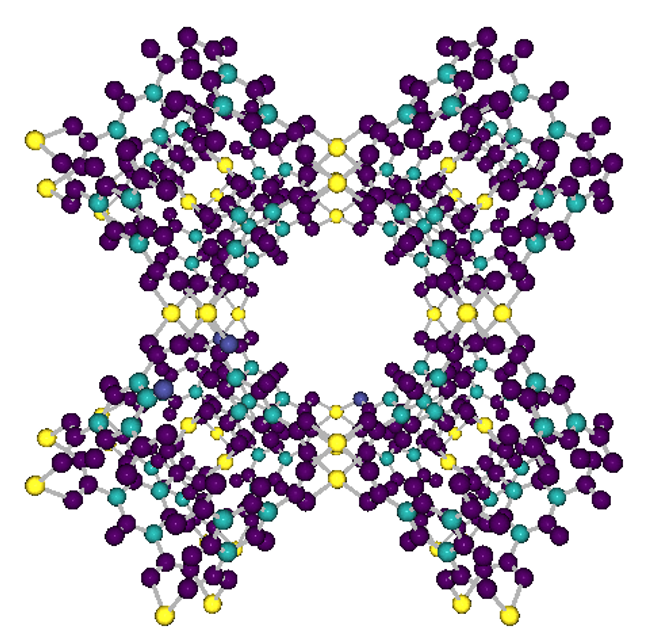}
        \caption{}
        \label{fig:hkust_anchor}
    \end{subfigure}
    \caption{Anchors on HKUST-1. (a) Crystal structure of the activated framework (Cu orange, O red, C gray, H white). (b) The same unit cell viewed along the $c$ axis with first-level anchors highlighted: Cu (yellow), carbon anchors (cyan; 95 of 96 are carboxylate-bearing ring carbons), O/H (blue); non-anchor atoms are purple.}
    \label{fig:hkust}
\end{figure}

Figure~\ref{fig:hkust} shows the crystal structure and the nodes selected as anchors. Only atoms inside one unit cell are drawn in Figure~\ref{fig:hkust_base}, so paddlewheels at the cell boundary appear without their ligands, whereas Figure~\ref{fig:hkust_anchor} completes such fragments; the graph itself uses periodic neighbors, so the model sees complete coordination environments. Because symmetry-equivalent atoms have identical local environments, they receive nearly identical importance scores and are selected or rejected together, as for the 48 Cu atoms and 95 of the 96 carboxylate-bearing carbons in seed 42. The anchor count is thus determined in units of symmetry classes ($48+96$), which is why the anchor fraction of this highly symmetric structure (23--24\%) exceeds the hMOF average of about 15\%. When a class lies close to the threshold, small numerical differences decide, which is the origin of the partial selections in Table~\ref{tab:hkust_seeds}, such as 40 of the 48 Cu atoms in seed 43. A fixed budget cannot reproduce this selection: top-$K$ with $K=32$ must pick 32 of the 48 equivalent Cu atoms, and a fixed ratio of 0.15 (94 anchors) must split the 96-atom carbon class. This case study illustrates why the number of clusters must be determined by the graph itself rather than fixed in advance.

\section{Gradient alignment and supervision scheme}
\label{sec:appendix_gradient_supervision}

\subsection{Training dynamics of gradient alignment}
\label{sec:appendix_gradient_alignment}

\begin{figure}[ht]
    \centering
    \begin{subfigure}[t]{0.48\linewidth}
        \centering
        \includegraphics[width=\linewidth]{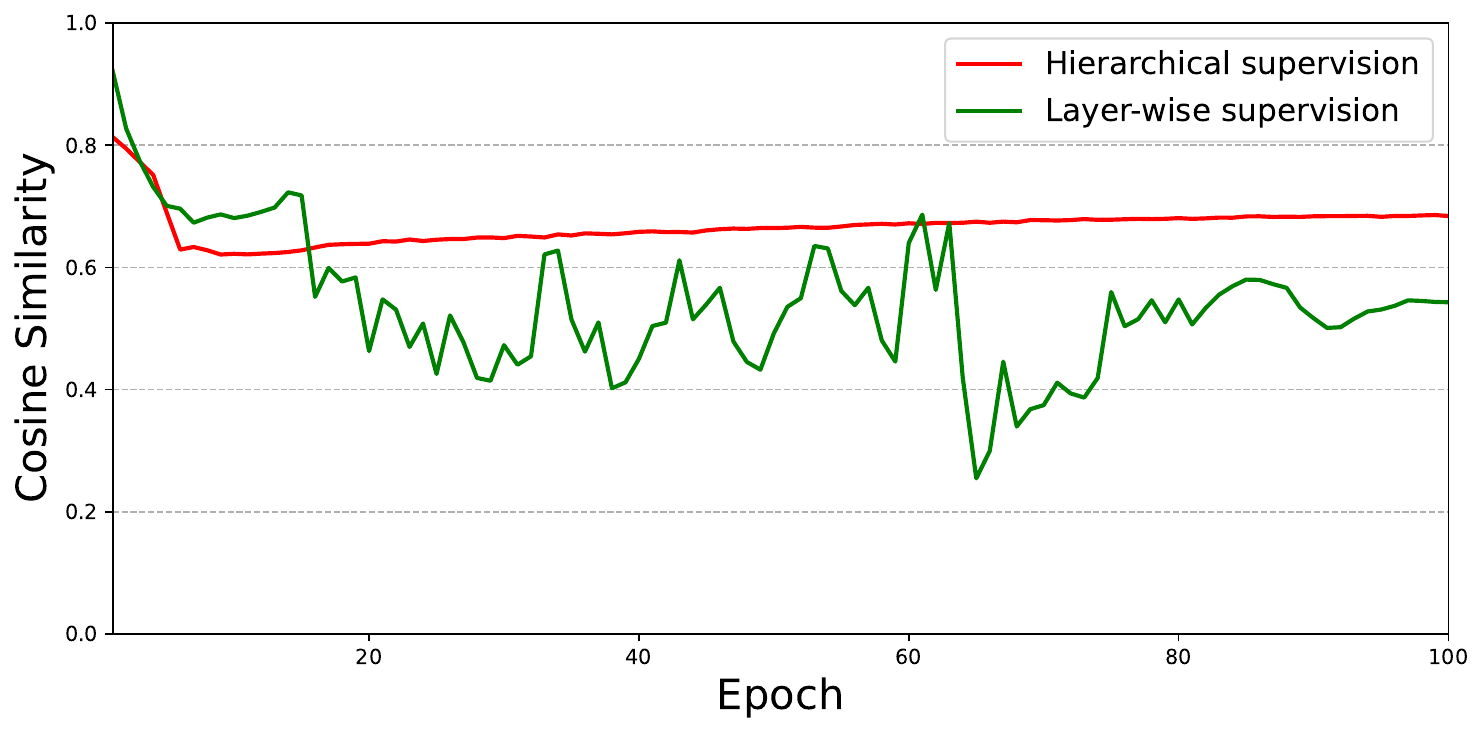}
        \caption{}
        \label{fig:gradient_similarity_gdpool}
    \end{subfigure}%
    \begin{subfigure}[t]{0.48\linewidth}
        \centering
        \includegraphics[width=\linewidth]{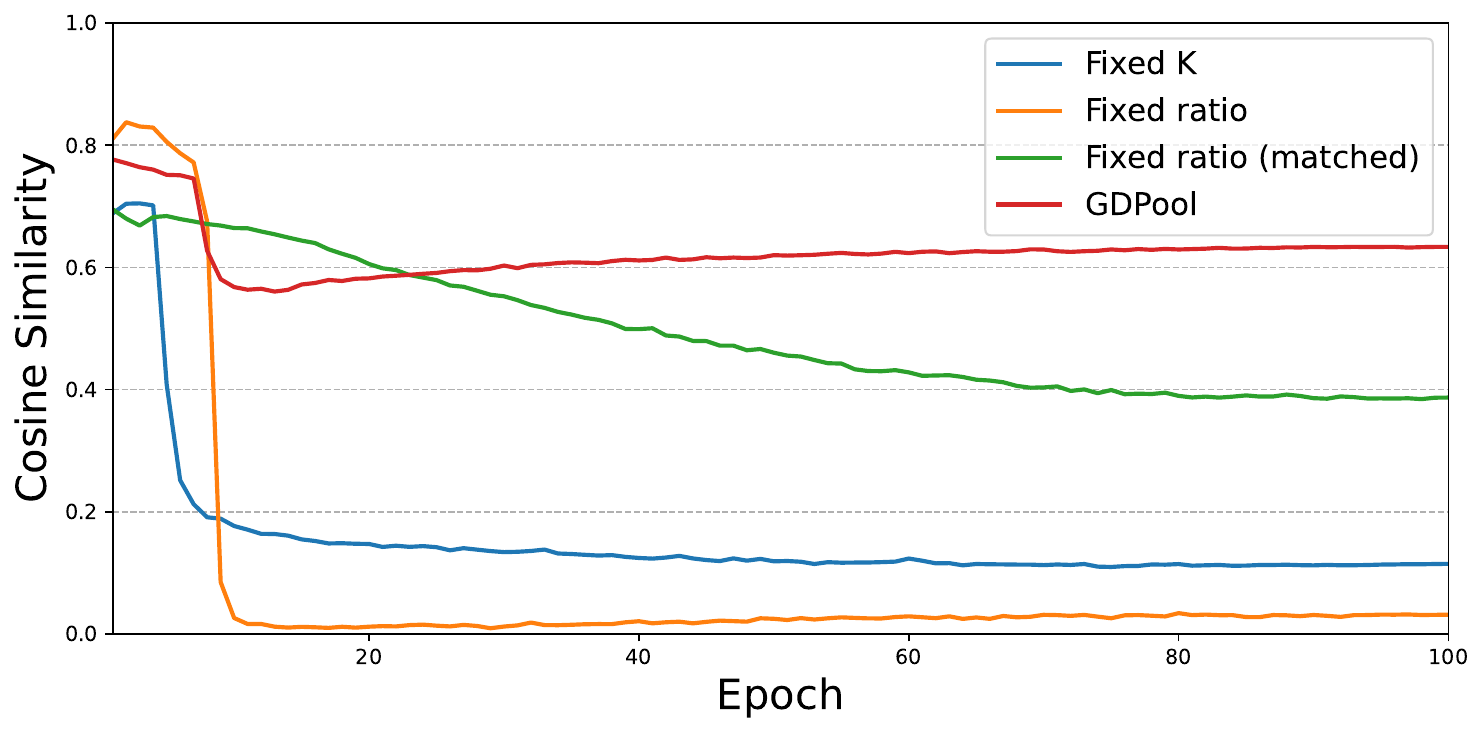}
        \caption{}
        \label{fig:gradient_similarity_as}
    \end{subfigure}
    \caption{Training-time gradient alignment across pooling levels. (a) Hierarchical vs.\ layer-wise supervision on the ARC-MOF dataset. (b) Anchor selection rules under hierarchical supervision for CO$_2$@0.1 on the hMOF dataset.}
    \label{fig:training_dynamics_gradient_alignment}
\end{figure}

This section provides results for the gradient alignment presented in Section~\ref{sec:Gradient_Alignment_Analysis}. The training dynamics are reported in Figure~\ref{fig:gradient_similarity_gdpool}. Final-level supervision has a single auxiliary objective, so inter-level alignment is undefined and it is omitted from Figure~\ref{fig:gradient_similarity_gdpool}. Under layer-wise supervision, the alignment of GDPool remains positive but lower and unstable; we relate this to the resulting hierarchy in Appendix~\ref{sec:appendix_supervision}. The anchor selection rule also affects gradient alignment. Figure~\ref{fig:gradient_similarity_as} compares the rules of Table~\ref{tab:anchor_rule} under the same hierarchical supervision. For the default fixed rules, the alignment between level objectives collapses within the first ten epochs, to about 0.1 for the fixed $K$ and near zero for the ratio of 0.5. The matched ratio decays throughout training to about 0.4, whereas GDPool remains at about 0.6. The resulting order of the four rules coincides with the order of their first-level activation rates in Table~\ref{tab:anchor_rule}. A plausible reason is that fixed rules impose the number of first-level anchors regardless of the graph, so some allocated anchors attract almost no nodes and form nearly empty clusters. The objective at the next level must nevertheless coarsen these clusters, which lowers its agreement with the first-level objective. The two measures thus point, structurally and in optimization, to the same limitation of fixed rules in forming active clusters.

\subsection{Effect of the supervision scheme}
\label{sec:appendix_supervision}

Hierarchical supervision differs from layer-wise supervision only in the arguments of the objectives: at level $l$, the former evaluates $\mathcal{L}_{\mathrm{Link}}$ and $\mathcal{L}_{\mathrm{Ent}}$ on the accumulated assignment $S^{(l)}_{\mathrm{Hier}}$ against the original adjacency $A^{(0)}$, the latter on the local assignment $S^{(l)}$ against the coarsened adjacency $A^{(l-1)}$. The two schemes share the first-level term, and the substitution above it changes the optimization in two ways. First, the reference graph becomes fixed. Under layer-wise supervision, $A^{(l-1)}$ is produced by the lower levels and moves during training, and each objective requires only that the pooled graph and the assignment agree with each other: a near-uniform assignment yields an almost constant $A^{(l)}$ that another near-uniform assignment matches up to scale, so no level is required to follow the input structure, consistent with the near-uniform assignments under layer-wise supervision in Table~\ref{tab:supervision}. Under hierarchical supervision, every level is compared with the same data-defined graph $A^{(0)}$. Second, the objectives become coupled: since $S^{(l)}_{\mathrm{Hier}} = S^{(1)} \cdots S^{(l)}$, each assignment matrix receives gradients from every objective at or above its level, all measured against the same graph, which is reflected in the higher and more stable alignment in Figure~\ref{fig:gradient_similarity_gdpool}. Final-level supervision keeps only the $l=L$ term of Eq.~\ref{eq:total_hierarchical_supervision}. Table~\ref{tab:supervision} compares the three schemes. Final-level supervision matches hierarchical supervision in accuracy while still outperforming all baselines, so the predictive advantage of GDPool stems from the operator rather than the supervision. The intermediate terms of Eq.~\ref{eq:total_hierarchical_supervision} instead shape the hierarchy: without them, only the final partition must preserve the input structure, and the first level retains more anchors with lower modularity. Layer-wise supervision degrades both accuracy and structure. Under layer-wise supervision, whose alignment is positive but unstable in Figure~\ref{fig:gradient_similarity_gdpool}, the first-level $K_\mathrm{eff}$ of GDPool nearly equals the number of anchors (65.3 of 65.8), yet its modularity collapses to zero in Table~\ref{tab:supervision}. This indicates a near-uniform assignment that trivially maximizes $K_\mathrm{eff}$ rather than clusters that follow the graph structure. Across supervision schemes and anchor rules, reduced or unstable alignment thus co-occurs with a failure in one of the two properties required for downstream analysis (Appendix~\ref{sec:appendix_anchor_rule}): fixed rules leave first-level anchors inactive, whereas layer-wise supervision yields assignments that do not follow the graph structure. Only GDPool with hierarchical supervision attains both high activation and high modularity, together with high and stable alignment.
\begin{table}[ht]
\caption{Effect of the supervision scheme on GDPool (ARC-MOF, CO$_2$@0.15). \#Anchors, modularity, and $K_\mathrm{eff}$ are reported per pooling level (L1\,/\,L2\,/\,L3).}
\label{tab:supervision}
\centering
\small
\begin{tabular}{lcccc}
\toprule
\textbf{Supervision} & \textbf{MAE} & \textbf{\#Anchors} & \textbf{Modularity} & $\boldsymbol{K}_{\mathrm{eff}}$ \\
\midrule
Layer-wise   & 0.250$\pm$0.007 & 65.8 / 38.3 / 5.6  & 0.00 / 0.00 / 0.00 & 65.3 / 38.2 / 4.9 \\
Final-level  & 0.152$\pm$0.001 & 53.7 / 24.7 / 12.1 & 0.16 / 0.32 / 0.35 & 49.5 / 23.0 / 11.5 \\
Hierarchical (GDPool) & \textbf{0.151}$\pm$0.000 & 20.7 / 10.5 / 5.9 & 0.42 / 0.41 / 0.23 & 18.5 / 10.0 / 5.8 \\
\bottomrule
\end{tabular}
\end{table}

\subsection{Hierarchical supervision with a different assignment operator}
\label{sec:appendix_hs}

We apply the same hierarchical supervision (Eq.~\ref{eq:total_hierarchical_supervision}) to DiffPool, whose auxiliary objectives are identical to ours, so that the comparison isolates the assignment operator. The architecture is identical to that of DiffPool in Table~\ref{tab:performance_comparison}, so this variant differs from DiffPool only in the supervision and from GDPool only in the assignment operator. Other baselines use different objectives, which would confound the operator with the loss. Both methods are trained with $\lambda \in \{0.05, 0.2\}$ under the protocol of Appendix~\ref{sec:appendix_model_configurations} for CO$_2$@0.1 on the hMOF dataset. Table~\ref{tab:hs} shows Hierarchical supervision with different assignment operators. At the same $\lambda$, GDPool achieves lower MAE than DiffPool (0.125 vs.\ 0.130 at $\lambda = 0.05$; 0.127 vs.\ 0.138 at $\lambda = 0.2$), and its accuracy varies less with $\lambda$. The two operators also differ in how the allocated clusters are used. DiffPool activates only 33\% and 21\% of its 32 first-level clusters at $\lambda = 0.05$ and $0.2$, respectively, whereas GDPool activates about 88\% of its first-level anchors at both values, even though the number of anchors changes from 19.7 to 33.2. $K_\mathrm{eff}$ of GDPool decreases monotonically with depth at both values, whereas that of DiffPool barely decreases from the first to the second level at $\lambda = 0.05$ (10.6 to 9.8) and increases at $\lambda = 0.2$ (6.7 to 12.7), so its hierarchy does not coarsen progressively. DiffPool attains comparable or higher modularity, particularly at $\lambda = 0.2$, but with most of its first-level clusters inactive, these clusters cannot support cluster-level analysis (Appendix~\ref{sec:appendix_anchor_rule}). Under identical supervision, the assignment operator thus determines whether the allocated clusters are used and whether the hierarchy coarsens progressively.
\begin{table}[ht]
\caption{Hierarchical supervision with different assignment operators for CO$_2$@0.1 on the hMOF dataset. \#Clusters, modularity, and $K_\mathrm{eff}$ are reported per pooling level (L1\,/\,L2\,/\,L3).}
\label{tab:hs}
\centering
\small
\begin{tabular}{lccccc}
\toprule
\textbf{Method} & $\boldsymbol{\lambda}$ & \textbf{MAE} & \textbf{\#Clusters} & \textbf{Modularity} & $\boldsymbol{K}_{\mathrm{eff}}$ \\
\midrule
DiffPool (Hier.) & 0.05 & 0.130$\pm$0.002 & 31.9 / 16.0 / 8.0 & 0.33 / 0.34 / 0.35 & 10.6 / 9.8 / 7.5 \\
DiffPool (Hier.) & 0.2 & 0.138$\pm$0.001 & 31.9 / 16.0 / 8.0 & 0.43 / 0.43 / 0.43 & 6.7 / 12.7 / 7.2 \\
GDPool & 0.05 & \textbf{0.125}$\pm$0.002 & 19.7 / 10.0 / 5.7 & 0.39 / 0.38 / 0.22 & 17.4 / 9.6 / 5.5 \\
GDPool & 0.2 & 0.127$\pm$0.001 & 33.2 / 8.9 / 4.5 & 0.25 / 0.35 / 0.27 & 29.4 / 8.2 / 4.3 \\
\bottomrule
\end{tabular}
\end{table}

\section{Sensitivity analysis results}
\label{sec:appendix_sensitivity_analysis}

This section provides results for the sensitivity analysis presented in Section~\ref{sec:Sensitivity_to_Model_Components}. The results show that GDPool maintains stable performance across a range of hyperparameter and architectural configurations, demonstrating its robustness.
\begin{table}[ht]
\caption{Effect of the entropy loss weight $\lambda$ on the ZINC dataset.}
\label{tab:Effect_lambda}
\begin{center}
\begin{tabular}{cccccc}
\toprule
\multirow{2}{*}{\boldsymbol{$\lambda$}} & \multirow{2}{*}{\textbf{MAE}} & \multicolumn{2}{c}{\textbf{Modularity}} & \multicolumn{2}{c}{$\boldsymbol{K}_{\mathrm{eff}}$} \\
\cmidrule(lr){3-4}\cmidrule(lr){5-6} 
& & Layer 1 & Layer 2 & Layer 1 & Layer 2 \\
\midrule
0.00
& $\textbf{0.239} {\scriptstyle \pm 0.003}$ 
& $0.208 {\scriptstyle \pm 0.003}$ 
& $0.095 {\scriptstyle \pm 0.004}$  
& $9.652 {\scriptstyle \pm 0.208}$ 
& $4.873 {\scriptstyle \pm 0.054}$ \\
0.05
& $0.240 {\scriptstyle \pm 0.015}$ 
& $0.283 {\scriptstyle \pm 0.002}$ 
& $0.168 {\scriptstyle \pm 0.014}$  
& $9.074 {\scriptstyle \pm 0.490}$ 
& $4.693 {\scriptstyle \pm 0.237}$ \\
0.10
& $0.241 {\scriptstyle \pm 0.010}$ 
& $0.351 {\scriptstyle \pm 0.005}$ 
& $0.237 {\scriptstyle \pm 0.010}$  
& $8.361 {\scriptstyle \pm 0.220}$ 
& $4.761 {\scriptstyle \pm 0.198}$ \\
0.15
& $0.240 {\scriptstyle \pm 0.009}$ 
& $0.374 {\scriptstyle \pm 0.011}$ 
& $0.372 {\scriptstyle \pm 0.011}$  
& $9.809 {\scriptstyle \pm 0.437}$ 
& $5.546 {\scriptstyle \pm 0.254}$ \\
0.20
& $0.251 {\scriptstyle \pm 0.009}$ 
& $0.370 {\scriptstyle \pm 0.002}$ 
& $0.449 {\scriptstyle \pm 0.003}$  
& $10.115 {\scriptstyle \pm 0.059}$ 
& $5.252 {\scriptstyle \pm 0.038}$ \\
\bottomrule
\end{tabular}
\end{center}
\end{table}

\begin{table}[ht]
\caption{Effect of pooling depth on performance and clustering behavior for the QM9 dataset.}
\label{tab:Effect_depth}
\begin{center}
\begin{tabular}{lcccccc}
\toprule
\multirow{2}{*}{\textbf{Method}} & \multicolumn{2}{c}{\textbf{MAE}} & \multicolumn{2}{c}{\textbf{Modularity}} & \multicolumn{2}{c}{$\boldsymbol{K}_{\mathrm{eff}}$} \\
\cmidrule(lr){2-3}\cmidrule(lr){4-5}\cmidrule(lr){6-7}
& 1$\times$Pooling & 2$\times$Pooling 
& 1$\times$Pooling & 2$\times$Pooling 
& 1$\times$Pooling & 2$\times$Pooling\\
\midrule
DiffPool
& $0.46 {\scriptstyle \pm 0.00}$ & $0.48 {\scriptstyle \pm 0.01}$
& $0.09 {\scriptstyle \pm 0.04}$ & $0.06 {\scriptstyle \pm 0.06}$
& $1.41 {\scriptstyle \pm 0.23}$ & $1.43 {\scriptstyle \pm 0.44}$ \\
MinCutPool
& $0.46 {\scriptstyle \pm 0.00}$ & $0.45 {\scriptstyle \pm 0.00}$
& $0.21 {\scriptstyle \pm 0.01}$ & $0.17 {\scriptstyle \pm 0.00}$
& $6.10 {\scriptstyle \pm 0.66}$ & $6.48 {\scriptstyle \pm 0.48}$ \\
MemPool
& $0.45 {\scriptstyle \pm 0.00}$ & $0.45 {\scriptstyle \pm 0.00}$
& $0.00 {\scriptstyle \pm 0.00}$ & $0.00 {\scriptstyle \pm 0.00}$
& $12.00 {\scriptstyle \pm 0.00}$ & $12.00 {\scriptstyle \pm 0.00}$ \\
BN-Pool
& $0.46 {\scriptstyle \pm 0.00}$ & $0.46 {\scriptstyle \pm 0.00}$
& $0.00 {\scriptstyle \pm 0.00}$ & $0.00 {\scriptstyle \pm 0.00}$
& $11.88 {\scriptstyle \pm 0.01}$ & $11.88 {\scriptstyle \pm 0.00}$ \\
TopKPool
& $0.45 {\scriptstyle \pm 0.00}$ & $0.47 {\scriptstyle \pm 0.01}$
& - & -
& - & - \\
SAGPool
& $0.45 {\scriptstyle \pm 0.01}$ & $0.49 {\scriptstyle \pm 0.01}$
& - & -
& - & - \\
ASAPool
& $0.47 {\scriptstyle \pm 0.00}$ & $0.50 {\scriptstyle \pm 0.00}$
& - & -
& - & - \\
MaxCutPool
& $0.56 {\scriptstyle \pm 0.09}$ & $0.67 {\scriptstyle \pm 0.09}$
& - & -
& - & - \\
EdgePool
& $0.45 {\scriptstyle \pm 0.00}$ & $0.45 {\scriptstyle \pm 0.00}$
& - & -
& - & - \\
\midrule
\textbf{GDPool}
& $\textbf{0.43}{\scriptstyle \pm 0.00}$ & $\textbf{0.40}{\scriptstyle \pm 0.00}$
& $0.25 {\scriptstyle \pm 0.00}$ & $0.23 {\scriptstyle \pm 0.01}$ 
& $6.71 {\scriptstyle \pm 0.08}$ & $6.78 {\scriptstyle \pm 0.51}$ \\
\bottomrule
\end{tabular}
\end{center}
\end{table}

\begin{table}[ht]
\caption{Sensitivity to the GFFN expansion ratio for CO$_2$@0.1 on the hMOF dataset.}
\label{tab:ffn_ratio}
\begin{center}
\begin{tabular}{ccclll}
\toprule
\textbf{GFFN Ratio} & \textbf{MAE}
& \textbf{Metric} & \textbf{Layer 1} & \textbf{Layer 2} & \textbf{Layer 3} \\
\midrule
\multirow{2}{*}{1}
& \multirow{2}{*}{$0.125 {\scriptstyle \pm 0.000}$}
& Modularity
& $0.384 {\scriptstyle \pm 0.002}$
& $0.382 {\scriptstyle \pm 0.001}$
& $0.211 {\scriptstyle \pm 0.002}$ \\
&
& $K_{\mathrm{eff}}$
& $17.759 {\scriptstyle \pm 0.183}$
& $9.685 {\scriptstyle \pm 0.079}$
& $5.493 {\scriptstyle \pm 0.090}$ \\
\midrule
\multirow{2}{*}{2}
& \multirow{2}{*}{$0.125 {\scriptstyle \pm 0.002}$}
& Modularity
& $0.388 {\scriptstyle \pm 0.003}$
& $0.384 {\scriptstyle \pm 0.005}$
& $0.217 {\scriptstyle \pm 0.006}$ \\
&
& $K_{\mathrm{eff}}$
& $17.418 {\scriptstyle \pm 0.305}$
& $9.595 {\scriptstyle \pm 0.176}$
& $5.534 {\scriptstyle \pm 0.070}$ \\
\midrule
\multirow{2}{*}{4}
& \multirow{2}{*}{$0.125 {\scriptstyle \pm 0.000}$}
& Modularity
& $0.390 {\scriptstyle \pm 0.003}$
& $0.381 {\scriptstyle \pm 0.002}$
& $0.218 {\scriptstyle \pm 0.002}$ \\
&
& $K_{\mathrm{eff}}$
& $17.166 {\scriptstyle \pm 0.223}$
& $9.565 {\scriptstyle \pm 0.034}$
& $5.504 {\scriptstyle \pm 0.050}$ \\
\midrule
\multirow{2}{*}{8}
& \multirow{2}{*}{$0.126 {\scriptstyle \pm 0.001}$}
& Modularity
& $0.392 {\scriptstyle \pm 0.003}$
& $0.382 {\scriptstyle \pm 0.003}$
& $0.226 {\scriptstyle \pm 0.006}$ \\
&
& $K_{\mathrm{eff}}$
& $16.921 {\scriptstyle \pm 0.257}$
& $9.668 {\scriptstyle \pm 0.194}$
& $5.658 {\scriptstyle \pm 0.129}$ \\
\bottomrule
\end{tabular}
\end{center}
\end{table}

\end{document}